\documentclass[article]{IEEEtran}
\usepackage{cite}
\usepackage[english]{babel}
\usepackage[utf8x]{inputenc}
\usepackage{graphicx}
\usepackage{xurl}
\usepackage{csquotes}
\usepackage{fancyhdr}
\usepackage{multicol}
\usepackage{etoolbox}
\usepackage{subcaption}
\usepackage{pdflscape}
\usepackage{hyperref} 
\usepackage{amsmath}
\usepackage{afterpage}
\usepackage{hyperref}
\usepackage{url}
\usepackage{array}
\usepackage{amsmath}

\begin{document}

\title{A Comparison of LLM Fine-tuning Methods and Evaluation Metrics with Travel Chatbot Use Case}

\author{
\IEEEauthorblockN{Sonia Meyer, Shreya Singh, Bertha Tam, Christopher Ton, Angel Ren}

\IEEEauthorblockA{Department of Applied Data Science, San Jose State University, San Jose, USA \\
Email: \{sonia.meyer, shreya.singh, bertha.tam, christopher.ton, angel.ren\}@sjsu.edu}
}

\maketitle

\begin{abstract}
This research compares large language model (LLM) fine-tuning methods, including Quantized Low Rank Adapter (QLoRA), Retrieval Augmented fine-tuning (RAFT), and Reinforcement Learning from Human Feedback (RLHF), and additionally compared LLM evaluation methods including End to End (E2E) benchmark method of “Golden Answers”, traditional natural language processing (NLP) metrics, RAG Assessment (Ragas), OpenAI GPT-4 evaluation metrics, and human evaluation, using the travel chatbot use case. The travel dataset was sourced from the the Reddit API by requesting posts from travel-related subreddits to get travel-related conversation prompts and personalized travel experiences, and augmented for each fine-tuning method. We used two pretrained LLMs utilized for fine-tuning research: LLaMa 2  7B, and Mistral  7B. QLoRA and RAFT are applied to the two pretrained models. The inferences from these models are extensively evaluated against the aforementioned metrics. The best model according to human evaluation and some GPT-4 metrics was Mistral RAFT, so this underwent a Reinforcement Learning from Human Feedback (RLHF) training pipeline, and ultimately was evaluated as the best model. Our main findings are that: 1) quantitative and Ragas metrics do not align with human evaluation, 2) Open AI GPT-4 evaluation most aligns with human evaluation, 3) it is essential to keep humans in the loop for evaluation because, 4) traditional NLP metrics insufficient, 5) Mistral generally outperformed LLaMa, 6) RAFT outperforms QLoRA, but still needs postprocessing, 7) RLHF improves model performance significantly. Next steps include improving data quality, increasing data quantity, exploring RAG methods, and focusing data collection on a specific city, which would improve data quality by narrowing the focus, while creating a useful product.
\end{abstract}


\section{Introduction}
The tourism and travel industry took a nosedive during the COVID-19 pandemic, which was declared by the World Health Organization (WHO) on March, 11, 2020, due to the severe lockdown measures restricting travel.\cite{nayak2022} Now, post-COVID, travelers are coming back with a vengeance and so was the “revenge travel” phenomena named to reflect the negative aspects of tourists exceeding the carrying capacity of destinations and business and governments desperately scrambling to recover income and grow the economy once again at the cost of tourism quality and sustainability.\cite{nguwi2022} The travel and tourism industry is projected to grow by 5.8\% per year for the next decade, which outpaces the overall economy at 2.7\%.\cite{binggeli2023}

The industry has faced challenges with widespread labor shortages due in short to poor working conditions, which along with the rise of large language model (LLM) applications presents a unique opportunity to incorporate technology into the travel and tourism industry.\cite{binggeli2023} Already, travelers overwhelmingly prefer to utilize technology for the full stack of travel from planning to booking to implementation. 74\% of travelers use the internet for travel planning, 45\% specifically use smartphones, 36\% are willing to pay more for an interactive and smooth booking experience, and 80\% prefer to use self-service to find information.\cite{peranzo2019} Technology is institutionalized with the global distribution system (GDS), an international reservation system for travel agents and suppliers, which is essential to online travel booking and customer service applications.\cite{peranzo2019}

Large language models (LLMs) were made possible with the 2017 “Attention is All You Need”\cite{Vaswani2017} transformer architecture without recurrent networks or convolutions, and work by predicting masked words or upcoming words.\cite{pahune2023} With LLMs, more data results in better predictions, most models have at least one billion parameters.\cite{pahune2023} They have been used for many purposes recently including carrying out task-based instructions, multilingual processing, medical and clinical applications, and programming.\cite{pahune2023} According to the literature, LLMs are typically trained on a huge corpus of text data with hundreds of billions of hyperparameters.\cite{zhao2023} Just in this year, 2023, we have seen several incredible applications utilize powerful LLMs, including in the booming travel industry. With the projected growth of the travel industry post-COVID and recent technological advances, there are potent opportunities for groundbreaking innovation with tangible effects on tourism income. Current artificial intelligence (AI) applications in travel focus on booking, however, there is a gap when it comes to artificial intelligence (AI) assisted travel planning and recommendations (as of May 2024). While not specifically tuned for marketing, ChatGPT remains a competitive chatbot for this application with its high quality conversational ability, however, to get customized travel recommendations, it requires very specific and detailed prompts, and it has a knowledge cut off of September 2021.

\begin{figure}[!h]
\centering
\includegraphics[width=.48\textwidth]{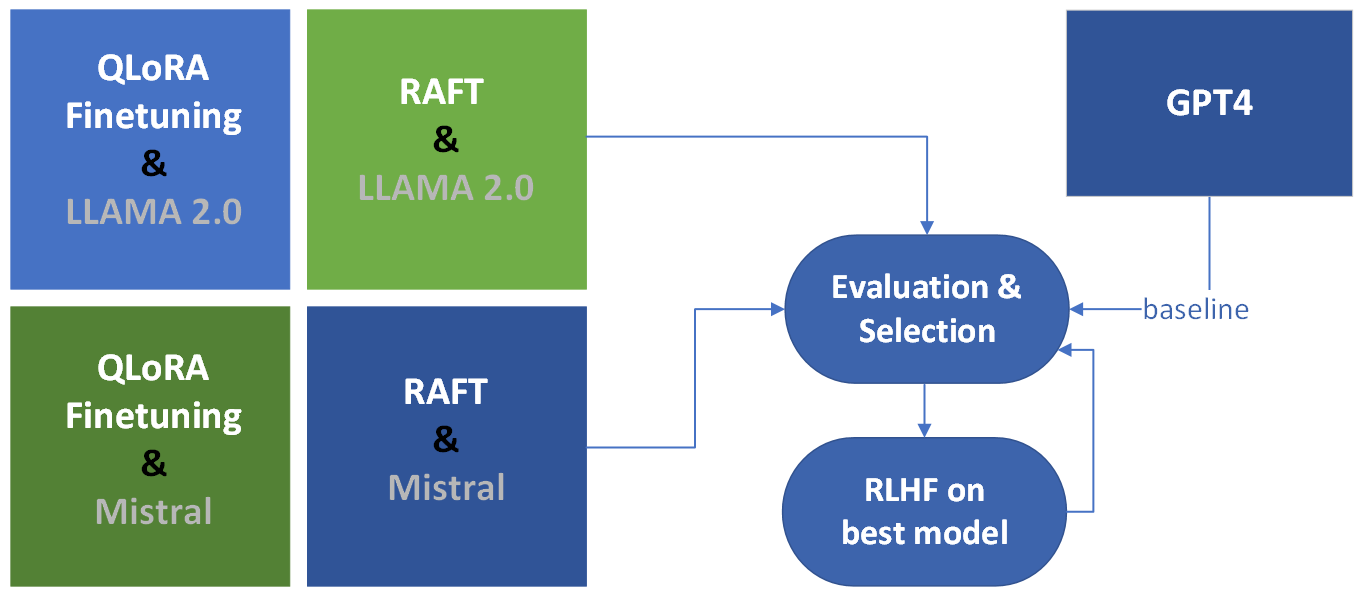}
\caption{Project Plan}
\label{fig:overview_init}
\end{figure}

Using the travel use case, this research compared two fine-tuning methods: 1) Quantized Low Rank Adapters (QLoRA) and 2) Retrieval-Augmented fine-tuning (RAFT). Two pretrained 7B models, LLaMa 2 and Mistral, are fine-tuned with these two methods, resulting in four models, then their inferences are evaluated against an extensive set of metrics using GPT as a baseline for comparison. The best model is fine-tuned with the third method, 3) Reinforcement Learning from Human Feedback (RLHF), resulting in 5 total models (see Figure \ref{fig:overview_init}). The evaluation metrics include: End to End (E2E) benchmark method of “Golden Answers”, traditional natural language processing (NLP) metrics, RAG Assessment (Ragas), OpenAI GPT-4 evaluation metrics, and human evaluation.

A highly effective fine-tuning technique called QLoRA was introduced in 2023.\cite{dettmers2023} Given the significant computational resources required for fine-tuning Large Language Models (LLM), QLoRA implements several innovative approaches to save memory without compromising performances. It applies gradient backpropagation through a frozen 4-bit quantized pretrained language model into Low Rank Adapters (LoRA). Additionally, it employs double quantization to lower the average memory demand. Furthermore, it incorporates paged optimizers to handle memory spikes. The second method implemented was a novel method called Retrieval Augmented Fine Tuning (RAFT), a training procedure for domain-specific Retrieval Augmented Generation (RAG), which can adapt pretrained LLMs like LLaMa 2 and Mistral for RAG in specific domains, such as ours in travel.\cite{zhang2024raft} RAG is a text generation method for outsourcing relevant information, from a knowledge base, or a large corpus of relevant, factual and quality information, to supply an LLM with contextual clues for producing factual responses. Then, an RLHF training pipeline will be done for domain-specific LLM curation, aligned with human preferences, with reward model training.\cite{iyer2023}

\section{Related Work}
Through a comprehensive review of literature and existing research papers, we build an understanding for state-or-the-art techniques and approaches aimed to achieve competitive performances. New innovations are expressed in different variations for enhancing large language models. Table \ref{table:lit_rev_comp} summarizes the selection of models, objective addressed, unique approach, performance results and ultimate findings. A survey of existing solutions show that initiatives have been developed to overcome shortcoming identified with public general-purpose and pretrained LLM. Some of the feature performance issues garnered from literature review are potential risks of hallucination, underdeveloped retrieval approaches, and inefficiencies in the use of computational resources.

\begin{table*}[!h]
\caption{Comparison of Different Models}
\label{table:tech_solutions_table}
\centering
\renewcommand{\arraystretch}{1.3}

\begin{tabular}{|  p{1.1cm}  |p{4cm}  |p{4cm}  |p{4cm}  |p{4cm} |} \hline 
 
 \textbf{ Models } & \textbf{Objective}	& \textbf{Approach} & \textbf{Best Results} & \textbf{Findings} \\ \hline 
 
 RAFT\cite{zhang2024raft} &
 Implement a novel approach called RAFT for domain-specific RAG, which addresses the naive nature of LLMs that are not able to distinguish between context and noise. &
 With this training dataset, they fine-tune LLM and use it to generate answers for given queries in specific domain. For each sample in training data all there is the question, answer (context), oracle documents (verified, relevant documents), and distractor documents. &
 They found the ideal combination of golden and distractor documents is a 1:4 ratio, thus 4 distractor documents to train the model to learn to extract relevant information and ignore these distractors utilizing chain of thought (CoT). &
 RAFT can enable smaller models like LLaMa 2 7B or Mistral 7B, to save inference cost and time. It is particularly useful for performing well in a specific domain, given a set of private enterprise information with specified knowledge. \\ \hline

 RAG\cite{semnani2023} & Current chatbots were not able to discuss niche topics and tend to generate inaccurate texts that sounded true, therefore spreading misinformation. & Retrieval augmented generation (RAG) was implemented to aid in response generation. Few-shot LLM based chatbot with reasoning and chain of thought prompting as well. & Proposed WikiChat outperforms all other chatbots based on niche topics discussed in conversation and compared metrics. & RAG improves accuracy while maintaining conversationality. \\ \hline 
 
 RAG\cite{piktus2021} & Information retrieval requires a hybrid fine-tuning recipe for RAG & Combines a parametric seq2seq model and dense representation model, effectively training a retriever and generator end-to-end. & RAG outperforms BART for generating diverse responses and answering Jeapordy questions. Its performance level is comparable to Robertafor fact verification. & RAG achieves state-of-art results without the need for extensive pre-training, eliminating need for re-ranker nor extractive reader. \\ \hline 
 
Generation-Augmented Retrieval (GAR)\cite{mao2021} & Augment and optimize user query through retrieval of relevant text, appending additional pieces to the original prompt. & Exploiting existing PLMs to generate relevant and supporting information to supply queries with additional context. & Extractive GAR outperforms other extractive methods for both NQ and Trivia datasets while generative GAR outperforms most of the other methods. & GAR benefits from using a sparse representation due to lightweight nature while remaining competitive in the generative space. \\ \hline 
 
 Retrieval-Augmented LLM\cite{liu2023} & A toolkit to build bespoke LLMs capable of enabling IR systems with LLMs. & The toolkit comprises 2 main modules for document retrieval and answer generation. The framework allows users options to reformulate their initial query, retrieve in-domain contexts from a knowledge base, and validate responses on a factual basis. & The toolkit is optimized for retrieval-augmented based LLMs and performance may not extrapolate for more general-purpose LLMs. & Contrast to LangChain, the customizable pipeline allows for the ingestion of various LLMs, build domain knowledge from documents and accurately link queries in a modularized way. \\ \hline 
 
 Fine-tuning with QLoRA\cite{dettmers2023} & Proposed three innovative approaches to save memory without compromising performance. & It applied gradient backpropagation through a frozen 4-bit quantized pretrained language model into Low Rank Adapters (LoRA). Additionally, it employs double quantization to lower the average memory demand. Furthermore, it incorporates paged optimizers to handle memory spikes. & Training on the Guanaco family of models, the second best model achieved 97.8 percent performance compared to ChatGPT on the Vicuna benchmark, trainable in under 12 hours on a single consumer GPU. The largest model reached 99.3 percent performance in just over 24 hours on a single professional GPU. The smallest Guanaco model (7B parameters) outperforms a 26 GB Alpaca model by over 20 percentage points on the Vicuna benchmark, while requiring only 5 GB of memory. & QLoRA can replicate 16-bit full fine-tuning performance with a 4-bit base model and Low-rank Adapters (LoRA). \\ \hline 
 
 Overview of LLMs\cite{zhao2023} & Provided a very in-depth overview of all the LLMS, their advantages, disadvantages. & A 97 page article surveying current LLMs, most currently updated as of September 2023. & LLaMa 2 is an open-source model that has around five times as many parameters as other models in the industry, but has shown high performance given instruction-based tasks. It is the latest open-sourced model that does well with instructional data that was released in July 2023. & Add retrieval-based practices during pre-training for the model to adapt and generalize well given new knowledge.
Storing training data within the parameters of a pretrained model is a challenge. There are parameter size constraints that need to be updated with new knowledge.\\ \hline
 
 \end{tabular}
\end{table*}

Zhang et. al introduced Retrieval Augmented fine-tuning (RAFT) as a novel training strategy for fine-tuning LLMs to better perform on RAG tasks. The key concept is data augmentation to generate "question, answer, document" triplets before fine-tuning. This is done by generating realistic questions paired with elaborate chain of thought answering scheme and purposefully including relevant and irrelevant context documents. Through a chain of thought with the distractor documents, the model learns to extract the correct information from the entire chunk of context through reasoning, ignoring the distractors. RAFT operates by training the model to disregard any retrieved documents that do not contribute to answering a given question, thereby eliminating distractions. The optimal ratio of oracle to distractor documents during training varies across datasets, but including some distractors improves generalization. Finally, during RAG, RAFT retrieves the top-k documents from the database. With RAFT, Zhang et. al addresses the limitations of fine-tuning with RAG, and evaluate their new methods on datasets including: PubMed, HotpotQA, and code documentation. RAFT consistently outperforms standard fine-tuning or RAG alone.\cite{zhang2024raft}

\begin{figure}[h]
\centering
\includegraphics[width=.48\textwidth]{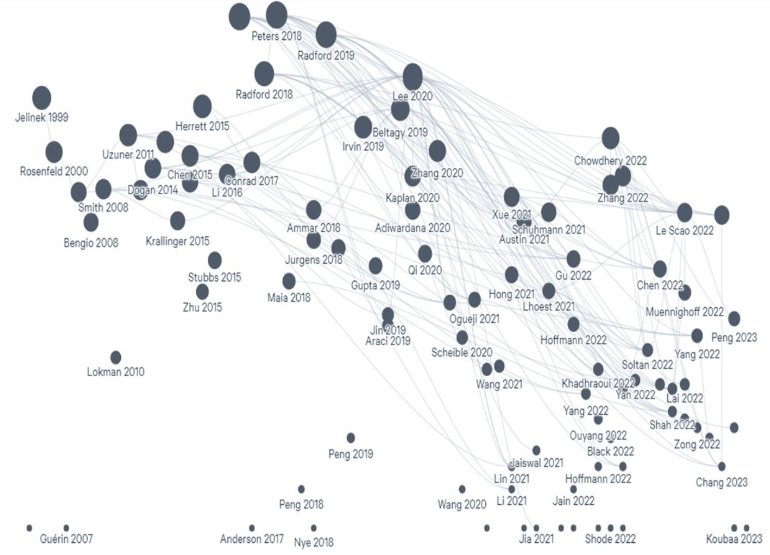}
\caption{\label{fig:llm_research_viz}Visualization of the explosion of LLM research from 2007 to 2023.\cite{pahune2023}}
\end{figure}

With the following literature survey on the existing research, we tried to gain an understanding on the current state of one of the most recent and fastest growing technologies (see Figure \ref{fig:llm_research_viz}). We attempted to get the basic understanding of the LLMs, different types of LLM chatbots, technologies, approach used, learnt different types of approaches works better in certain types of research works, and the performance of these may vary depending on the tasks. Some of the research studies shows, some chatbots perform well without the fine-tuning as well. Below is the summary of different research papers the team has gone through, some of the papers are summarized in the tabular form as well (see Table \ref{table:lit_rev_comp}), showcasing the brief on the goal, approach used and its conclusion. These papers helped the team to have a better understanding on the topic.

Gudibande et al. (2023) warns against the temptation of open source language models trained on imitation data to mimic ChatGPT’s proprietary abilities.\cite{gudibande2023} These models use ChatGPT conversations with real users collected through ShareGPT, Human-ChatGPT Comparison Corpus, r/ChatGPT subreddit, and the Discord ChatGPT bot to train.\cite{gudibande2023} For reference, ChatGPT uses between 1.5 and 3 billion parameters depending on the model, while these imitation models use only 0.3-150 million parameters.\cite{gudibande2023} Such imitation models include Alpaca, Vicuna, Koala, and GPT-4ALL.\cite{gudibande2023} Despite comparable evaluation on crowd rating and canonical NLP benchmarks and even better performance on simple instructions, the imitation models simply excel at imitating ChatGPT deceiving crowd raters, but actually there is a big gap when it comes to accuracy and breaking down complex tasks.\cite{gudibande2023}

Zhou et al. (2023) introduces LIMA, a 65B parameter language model fine-tuned on 1,000 curated prompt-response pairs.\cite{zhou2023} The model was highly generalizable, and specifically mentions having trip planning ability. Ablation experiment showed diminishing returns with scaling quantity of data, just 30 dialogue examples dramatically improved multi-turn conversation ability, which was measured by the rate of excellent responses going from 45\% to 75\%.

Lin et al. (2023) propose LLM-EVAL, a unified automatics evaluation method for conversations using LLMs, which is much more simplified and efficient in comparison to current methods using human annotation, ground truth responses, and multiple LLM prompts. LLM-EVAL is a single prompt based evaluation method that uses a unified schema to evaluate conversational quality. Lin claimed that traditional evaluation metrics like BLEU and ROGUE are insufficient for natural conversations. LLM-EVAL outperformed other supervised, unsupervised, and LLM-based evaluation metrics. Supervised techniques include dialogue quality (fluency, relevance, knowledge) and topic transition known as GRADE. Unsupervised techniques include DEB, which is BERT fine-tuned with relevant and adversarial irrelevant responses, and FED, which uses embedding features and probabilities to determine dialogue quality. LLM based methods include GPTScore to assign quality ratings, using InstructGPT, and G-EVAL which uses chain of thought and a form filling paradigm. They use several chat datasets like DSTC10 Hidden Set, including JSALT, NVM, ESL, Topical-DSTC10, and Persona-DSTC10. They took Spearman correlation coefficient to compare the difference between human ratings and automatic metrics across appropriateness, content, grammar, and relevance, and additionally tested and compared the use of LLM-EVAL on different LLMs including Anthropic Claude and OpenAI ChatGPT. Claude and ChatGPT are optimized for chat, and performed better with LLM-EVAL.\cite{lin2023}

Zhao et al. (2023) surfaces the observation that as multimodal capabilities of generative algorithms become world apparent, their explicit knowledge is subjected to hallucination. World knowledge exists in the forms of images, visuals, and audio formats. Each format would require its own domain-specific retrieval-augmented generation method. Specifically for tasks that query for knowledge, there exists 2 main open-ended challenges. Conventionally, one of the challenges involves storing training data within the parameters of a pretrained model. Limitations to this approach include the size constraint of the parameters and the need to update the parameters with new knowledge. The second challenge is pairing a generative model with a retrieval step to achieve better responses from the input data or external references. A remedy for reducing hallucinations is a foundation of standardized knowledge index. However it is structured, this foundation can be transformed into a dense representation that is optimal for efficient storage and search.The recommendation is to integrate retrieval-based practices into the pre-training process so that the generative model can better generalize against new knowledge and adapt accordingly.\cite{zhao2023}

Banerjee et al. (2023) found that LLM-powered chatbots are effective platforms to provide services to customers, but there is an important need to assess its performance in efficient and measurable ways. In order to benchmark a chatbot’s performance, accuracy and usefulness will serve as the two main characteristics for evaluation. The first being how accurate the chatbot is able to complete a set of tasks and the second being how useful it is able to fulfill a user’s needs. The proposed metrics is the End to End (E2E) benchmark method using cosine similarity given a set of predefined answers called the “Golden Answers.” The E2E metric compares the chatbot’s output results to an expert human answer, or the “Golden Answer.” They used the Universal Sentence Encoder (USE) and Sentence Transformer (ST) models to create vectorized embeddings and obtain the underlying context and meaning of the texts\cite{banerjee2023}. Banerjee et al. and Lin et al. argues that traditional metrics like Recall-Oriented Understudy for Gisting Evaluation (ROUGE), which uses n-grams overlaps, are insufficient to capture the deep complexity and semantic meaning of a conversational chatbot.\cite{banerjee2023}\cite{lin2023} E2E is user centric, considers semantic meaning, and improved with advanced prompt engineering alongside human evaluation metrics, unlike ROUGE.\cite{banerjee2023}

Maroengsit et al. (2019) discusses the difficulties to evaluate and compare different chatbot systems based on their effectiveness, efficiency, ability to satisfy users, and achieve their specific goal or desired prompt. Chatbots have been around since the 1960’s and there has been many technological changes to advance chatbot development. However, there has not been an updated holistic study on evaluation methods of chatbots and there’s a huge potential for chatbot developers to apply these evaluation techniques to their own work. There are two types of chatbots, rule-based and AI-based. Depending on the chatbot type there will be different evaluation methods used to asses those architectures. The main steps of architecture processes in chatbot systems are Natural Language Processing (NLP), Natural Language Understanding (NLU), and Natural Language Generation (NLG). While NLP focuses on text preprocessing via pattern matching, parsing, Term Frequency-Inverse Document Frequency (TF-IDF), or Word2Vec, NLU is processing by obtaining semantic understanding of texts. This is accomplished by having the model understand the conversation between the user and the model itself using the following techniques commonly found in the literature: intent classification, dialogue planning, name entity recognition, vector recognition using cosine similarity, lexicon, or long short-term memory. NLG is response generation, where the point is to gain information from a user’s response by asking them a question and seeing how they reply. Essentially, NLG determines how the system, or agent, responds to the user base from the information gathered in each conversation. There are three main evaluation methods: content evaluation, user satisfaction, and functional evaluation. Content evaluation uses automatic evaluation metrics such as precision and recall using BLEU or ROUGE. This is beneficial because even though having human judges for evaluation is accurate, it takes a long time and is expensive. Thus, it is more efficient and faster evaluation time with lower costs. User satisfaction is based on a Likert five-point scale, which uses human users’ satisfaction surveys to evaluate complex chatbot systems where there is not just one correct answer. However, the issue is the presence of bias. Future chatbot developers can use these results as a guideline when deciding the most well-suited evaluation methods based on their uses cases.\cite{maroengsit2019}

Svikhnushina et al. (2023) proposed a Dialog system Evaluation framework using Prompting (DEP) to specifically evaluate social chatbots with prompt engineering, where the prompts guide the process of how a chatbot generates a response. The issue is that prompt-based learning hasn’t been studied enough when evaluating dialog systems, so they focused on a comprehensive evaluation of dialog chatbot systems using prompting. There are three types of prompting, zero-shot, one-shot, and few-shot prompting. Zero-shot prompting of LLMs do not need previously labeled data to do new tasks, while few-shot prompting requires the LLMs to have some labeled prompts to conduct new tasks. The three evaluation scores they used were “Bad,” “Okay,” and “Good.” The DEP method essentially collects chat logs between the LLM and chatbot system and then prompts the LLM to generate scores based on the chat logs. However, there are several problems such as being unable to “capture subjective perceptions”.\cite{svikhnushina2023} when the user chats with the system, and likely is not able to generalize to real-world settings. Therefore, there is a need and importance to optimize social chatbots by streamlining the evaluation process in a more efficient manner. Their solution to streamline the evaluation process was to removing human involvement during each step, saving time and cost. Instead of using LLMs that use prompting, which tend to create misinformation, they gave instructions and prompted LLMs to conduct a specific behavior using the proposed DEP framework, which they found yielded better results. This indicates how you can leverage LLMs to create realistic conversations. Thus, they found that the most well-performing prompts contain few-shot learning and instructions, which show its ability to generalize on a corpora of data.\cite{svikhnushina2023} 

Irvine et al. (2023) focuses on utilizing human feedback and intuitive metrics to create engaging social chatbots to improve user retention and engagement based on human feedback. In order to aid chatbots in understanding the intentions of a user’s query, feedback was collected automatically generated pseudo-labels during every interaction with the user. These serve as user engagement proxies to train a reward model, so when the chatbot generates a poor sample response during inference time, or the time it takes to complete the forward propagation, then it will be given a low score. These are basically the worst responses that will be rejected because they do not make sense, are not appropriate responses, etc. Four types of metrics were used, such as mean conversation length (MCL) to measure the chatbot’s engagement level with a user, retry rate, user star rating, and retention rate. The retry rate is how often user requests to regenerate a response and the user star rating is essentially user satisfaction feedback based on a rating scale. On the other hand, retention rate is the most challenging and expensive evaluation rate. The problem is that it is manually extensive, costly, and time consuming to have any human involvement such as expert annotators rank responses. Although human experts can improve the responses that LLMs generate, these disadvantages may likely limit the chatbot development and evaluation process instead. Therefore, there’s a lack of engaging chatbots that retains users to come back and talk, so instead of purely fine-tuning LLMs on conversational data, one may incorporate human feedback during development to collect more data for the human reference or knowledge base, which has shown improved retention scores of over 30\%.\cite{irvine2023} Thus, the solution is to use human feedback to create engaging chatbots and improved engagement levels. This assumes that the longer a user chats with the system, the more likely they are to return and chat in the future, indicating higher retention. It was found that chatbots have longer average user interactions and user retention when grouped training reward models with the human feedback methodology.\cite{irvine2023}

Wong et al. (2023) demonstrate how LLMs, particularly ChatGPT, have the ability to drastically change the tourism industry, such as improving customer experience in 3 travel stages: before the trip, en-route, and post-trip encounters. Unlike previous AI, it is able to improve trip planning efficiency, create more personalized recommendations by having it act as a tour guide or local, and having a readily available and streamlined personal guide in your pocket during the trip in cases of emergency. ChatGPT provides efficient travel solutions by looking for travel information, filtering out irrelevant travel information, and then makes a comparison of different options in a neat fashion that is human-readable. The problem is when travelers research where to go using the internet, through travel agencies, or from word of mouth, there is a huge abundance of information. This makes it difficult, overwhelming, and time-consuming to search for valuable information and make decisions on where and what to go or do. There are a lot of communication problems when traveling such as a lack of cultural understanding, language barriers, etcetera. The largest limitation of utilizing LLMs such as ChatGPT for travel, is that it is still not entirely accurate since sources can be misleading. ChatGPT is unable to fully explain their answers in a rational way, bringing up transparency and accountability issues. The responses can be biased bringing up ethical issues and spreading disinformation. This is because there could be prejudiced information in the training data itself and there is limited domain knowledge as it is not able to be trained on certain semantics and ways of human expression, which could cause the LLM to misinterpret the user query. Also, the data is limited to the year 2021 so any information after that date is not in the training set at all. However, despite these limitations, one can utilize the benefits of using LLMs such as ChatGPT across the three stages for a tourist, before, during, and after traveling. These advantages and disadvantages of LLMs in the travel and tourism industry are important takeaways to take into consideration while developing our travel domain-specific chatbot.\cite{wong2023}

  \begin{table*}[!h]
    \caption{Literature Survey Comparison}
    \label{table:lit_rev_comp}
    \centering
    \begin{tabular}{|p{1.3cm}|p{4cm}|p{4cm}|p{4cm}|p{4cm}|}
      \hline
      \textbf{Authors} & \textbf{Goals} & \textbf{Approach} & \textbf{Targeted Problem} & \textbf{Conclusion} \\
      \hline
      Maroengsit et al. (2019)\cite{maroengsit2019} &
      Conduct a comprehensive study on evaluation methods of chatbots that has never been done before and has not been updated despite how long chatbots have been around for. & 
      The architecture is based on NLP, NLU, and NLG. There are three main evaluation methods: content evaluation, user satisfaction, and functional evaluations based on a task or goal, usage statistics, and building blocks. & 
      It is difficult to ensure that the automation of customer contact is of acceptable quality. There’s a lack of comprehensive studies on evaluation methods to evaluate chatbot systems overall. & 
      Results can serve as a guideline for future chatbot developers and help them decide what evaluation methods are best suited for their chatbots, based on the type (AI vs rule-based or both), domain (specific or general), goal, and purpose. \\
      \hline
      Caldarini et al. 2022\cite{caldarini2022} &
      Generate a literature survey on DL and NLP techniques used in most recent chatbots, challenges and limitations of implementing chatbots, and recommendations for future research. &
      Instead of using rule-based chatbots with limited capabilities, AI chatbots are more flexible, which have been trained through ML algorithms using a training dataset. &
      Relying on pattern matching rules cause chatbots to be domain dependent and inflexible, and understanding human language is challenging. &
      After surveying 59 articles for their study, they highlighted gaps in the literature and made a list of recommendations to help future chatbot developers before they get started, saving research time and money. \\
      \hline 
      Svikhnushina et al. 2023\cite{svikhnushina2023} &
      Prompt-based learning has not been studied enough in the literature so they proposed a DEP framework to evaluate dialogue chatbot systems using prompt engineering. &
      Their proposed DEP method collected chatlogs between the backend LLM and frontend chatbot systems. Then, they utilized the LLM to generate scores given previous chat logs. It was found that few-shot prompting performed better than zero-shot prompting. &
      The system is unable to “capture subjective perceptions” between the user chatting with the system and lacks the ability to generalize well on the data. Current literature is lacking the need to streamline an efficient evaluation process to optimize chatbot performance. &
      The solution is to create a streamlined evaluation process by removing human involvement during each step, saving time and cost. They also highlight how one can leverage LLMs to create realistic conversations. \\
      \hline
      Irvine et al. 2023\cite{irvine2023} & 
      Create socially engaging chatbot systems that can understand the intentions of a human user’s query using human feedback and intuitive metrics. Also, to improve user retention and engagement. & 
      Collect feedback using pseudo-labels that are generated during user interactions. Metrics to evaluate how a chatbot was able to improve user retention and engage with the user: mean conversation length (MCL), retry rate, user star rating, and retention rate. & 
      Chatbots are not engaging enough to keep user retention. Using reinforcement learning from human feedback can improve the responses that LLMs generate but are very costly and time-consuming, limiting chatbot development. & 
      The average user interaction and retention rate is higher when training reward models is grouped with the human feedback method. \\
      \hline
      Kang et al. 2023\cite{kang2023} &
      Replace the use of tour guides utilizing a smart tourism chatbot to help users find travel information, decide, and curate travel plans. &
      Chatbot system contains: gateway, the DST server, the Neo4J graph DB server, and the management server and additional features of map view and real-time weather data. Q\&A datasets were also developed using pretrained language models. & 
      The problem statement is not well-defined, and that the main issue is that the smart chatbot continues to generate some inaccurate answers due to the nature of the data. Thus, it’s important to assess the feasibility and practicality of a travel-specific smart chatbot with a well-defined problem statement. & 
      Develop a chatbot using a “tourism information multi-domain DST model and Neo4J graph DB to retrieve tourism information for context-aware personalized travel planner service” by utilizing transfer learning with a pre-defined multi-domain DST dataset containing tourism data. \\
      \hline
      Trivedi et al. 2022\cite{trivedi-2022} & 
      Existing LLMs struggle when the necessary knowledge is either unavailable to the LLM or not up-to-date within its parameters. Here, what to retrieve depends on what has already been derived, which in turn may depend on what was previously retrieved. IRCoT, a new approach for multi-step Q\&A that interleaves retrieval with steps (sentences) in a CoT, guiding the retrieval with CoT and in turn using retrieved results to improve CoT. &
      A new approach called IRCoT, which interleaves retrieval with chain-of-thought (CoT) reasoning. It uses the current CoT to guide the retrieval of additional passages from the external knowledge source. IRCoT then uses the retrieved passages to improve the CoT by adding new steps or modifying existing steps. This process continues until IRCoT is able to generate a complete and accurate CoT for the question. & 
      Existing LLMs struggle with the multi-level Q\&A because they are often prone to hallucination and factual errors when reasoning about complex topics or answering open-ended questions. & 
      Leveraged the ability of Chain of thought prompting that has significantly improved LLMs to perform well in multi Q\&A tasks. The authors have introduced IRCoT, which uses interleaved CoT reasoning and retrieval steps that guide each other step by step. \\
      \hline
      Cao et al. 2023\cite{cao-2023} & 
      To develop a chatbot that can better engage in professional conversations with users. The author emphasizes that existing LLMs are not well suited for task-oriented dialogue because they lack the ability to automatically manage topics and track dialogue state. &
      It uses a four-stage workflow to conduct task-oriented dialogue:
      1) topic development: generates list of topics to discuss, 2) maintaining topic stack: to track conversation state of previous/current topic, 3) enriching topic: retrieve relevant information from external knowledge sources, 4) generating response &
      Chatbots are better in dealing with task-oriented dialogue; chatbot must understand the user's request, track the conversation state, and generate relevant helpful responses. Existing LLMs are not well suited because they lack the ability to automatically manage topics and track dialogue state. This can lead to problems like the chatbot getting off track and repetitious or irrelevant responses. &
      The authors evaluated DiagGPT on a dataset of simulated user conversations and found that it outperformed existing methods on both dialogue quality and task completion rate. From this, we can infer that this approach can be helpful in developing chatbots that are better able to engage in task-oriented dialogue. \\
      \hline
      Wei et al. 2022\cite{wei-2022} & 
      Introduce chain-of-thought (CoT) prompting as a new method for improving the reasoning ability of large language models (LLMs). & 
      Input sequence of prompts/instruction to LLM, which generates text that completes each prompt. The output is used as the input for the next prompt. This process continues until the final prompt is reached, and the output of the final prompt is the LLM's response to the original query. & 
      It enables the LLM to perform better in the reasoning by providing some steps explicitly. & 
      It shows that CoT improves the performance of the LLMs when they trained on small datasets. \\
      \hline
    \end{tabular}
  \end{table*}

\section{Travel Dataset}

\begin{table}[!h]
\caption{A Review Summary of Models with Data Structure}
\label{table:model_with_data}
\renewcommand{\arraystretch}{1.3}
\begin{tabular}{|p{2cm}|p{5cm}|}
 \hline
\textbf{Model} & \textbf{Data Structure} \\
 \hline
Quantized Low Rank Adapter (QLoRA) & QLoRA is an effective, highly adaptable fine-tuning approach that needs structured Question-and-Answer (Q\&A) format.\cite{rao2023} \\
 \hline
Retrieval Augmented fine-tuning (RAFT) & In addition to Q\&A format, RAFT includes detailed chain of thought (CoT) for reasoning, verified relevant and irrelevant documents for distinguishing relevant context from noise.\cite{zhang2024raft} \\
 \hline
Reinforcement Learning from Human Feedback (RLHF) & The Q\&A data is typically already generated by an LLM, then evaluated by humans as good or bad. In this model, the top rated Reddit comments were good and lowest rated bad.\cite{schmid2024} \\
 \hline
\end{tabular}
\end{table}

\begin{figure*}[]
\centering
\includegraphics[width=.8\textwidth]{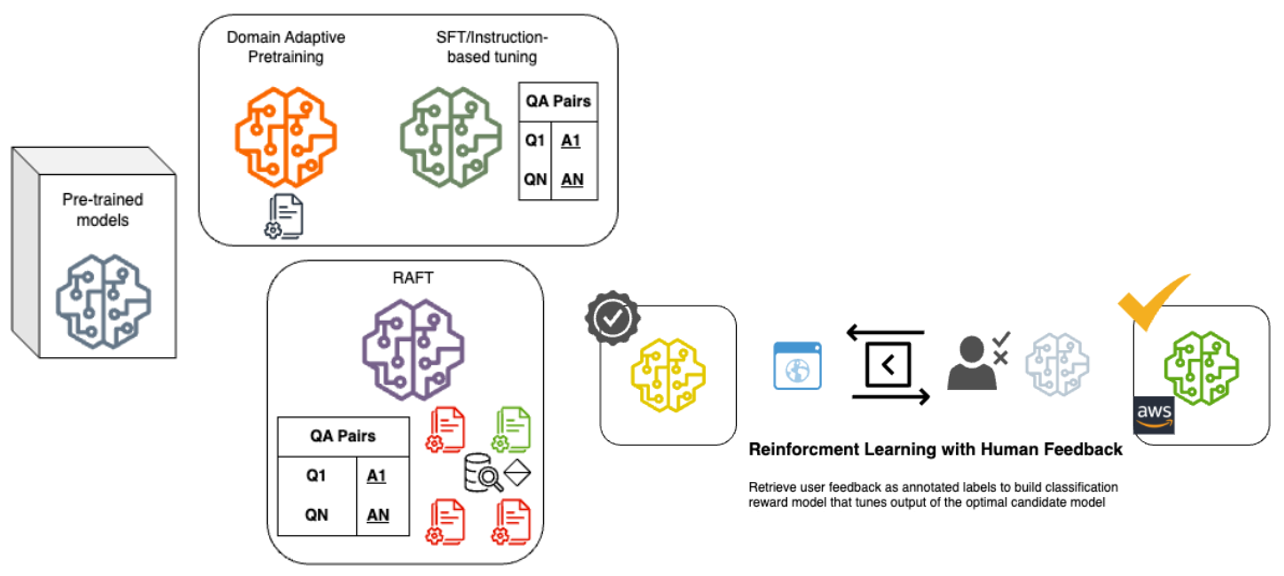}
\caption{\label{fig:5.3e2e}Project Overview and Data Structures}
\end{figure*}

Table \ref{table:model_with_data} provides a high level overview and Figure \ref{fig:5.3e2e} provides a visualization of the data structure needed for each model. The travel dataset was collected entirely from Reddit, and specifically pulled from travel domain subreddits. There are a plethora of travel subreddits available, with r/travel being the largest and most active with 8.9 million subscribers. It is one of the most popular communities in the top 1\% of subreddits as of December 2023.\cite{reddit_travel_subreddits} Calls made to the Reddit API to curate a corpus of Question-and-Answer (Q\&A) formatted data, collected conversational data from 201 subreddits.\cite{keita2023}  This consists of 27 travel-related subreddits, 30 country subreddits, and 144 city subreddits which include r/solotravel, r/travelhacks, r/roadtrip, and so on so the data can capture the context of conversations. The data collected for this project was for the purpose of fine-tuning the chatbot, providing domain specific knowledge, and to provide current updated travel information to address LLM knowledge cutoffs. Due to the Large Language Model (LLM) capabilities of generating recommendations due to its pretraining\cite{dai2023}, the Reddit post would serve as the question and the compiled comments as the answer. Tools used for data collection and processing include: Python, PRAW, Pandas, NLTK, Regex, BERTopic, and Ollama.

\subsection{Data Collection}
ChatGPT was also used to generate the top 30 countries for tourism, and tourist cities were collected from Wikipedia. A targeted list of travel subreddits was collected from a blog post, however, subreddits with a heavy focus on sharing images were omitted.\cite{reddit_travel_subreddits} In total, there were 201 subreddits: 27 travel related subreddits, 30 country subreddits, and 144 city subreddits. There are a total of 16,300 entries sourced from Reddit, which was further preprocessed and reduced the entries from 16,300 to 10,500 rows. 

\begin{figure}[!h]
\centering
\includegraphics[width=.48\textwidth]{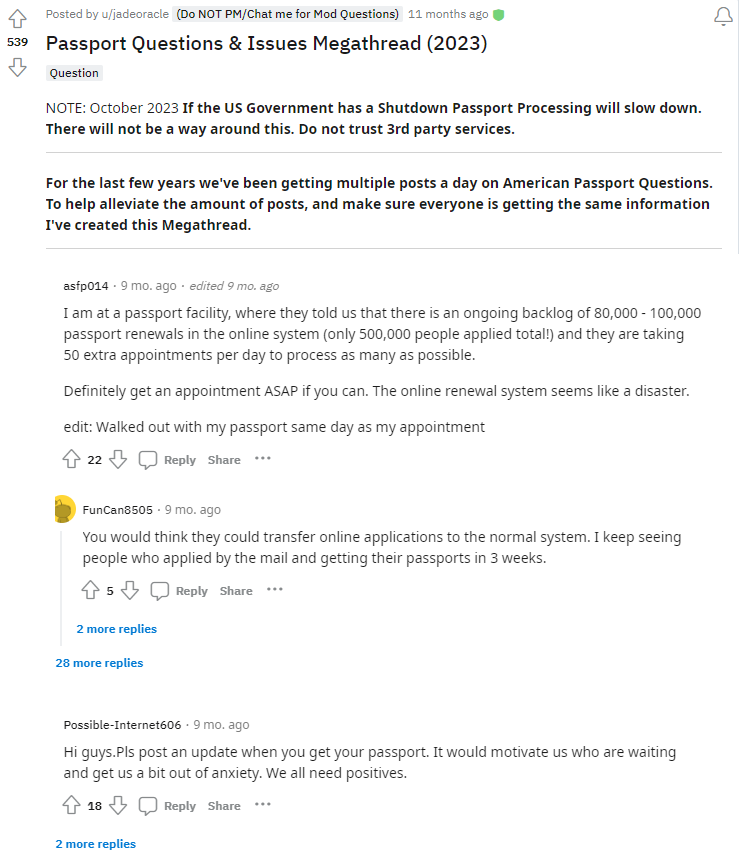}
\caption{\label{fig:reddit_travelpassport_post} Example of the Hottest Post Format from Reddit}
\end{figure}

Reddit data is appropriate for question and answer type data as each post has multiple comments or answers. Reddit is categorized by subreddits, which allow for convenient filtering for travel related questions and answers by targeting travel specific domains. Python Reddit API Wrapper (PRAW) was used to make API requests.\cite{praw} Two methods of collecting posts were used: 1) hot, and 2) top. Top means best of all time and takes a time frame as a parameter. Top 1,000 posts of the year in travel related subreddits, top 50 posts of the year in country subreddits, and top 20 posts of the year in city subreddits. Hot is what is currently trending, so these posts are collected daily with smaller requests to reduce requesting the same post multiple times if it is trending for multiple days. Hot 100 posts of travel related subreddits, and hot 20 in country and city subreddits. The top of the year posts were collected at once for the prior year, whereas the hot daily posts were collected daily for 4 days. Despite making so many requests, only 4,000+ posts were collected, perhaps due to not having enough posts within subreddits to meet the request 1,000 or some subreddits did not exist. Figure \ref{fig:reddit_travelpassport_post} is a snapshot of the hottest post on the travel subreddit titled “Passport Questions \& Issues Megathread (2023),” along with two sample comments from multiple Reddit users, along with details such as upvotes and the date it was posted.\cite{passport_reddit_post_comments} 

\begin{figure}[!h]
\centering
\includegraphics[width=.48\textwidth]{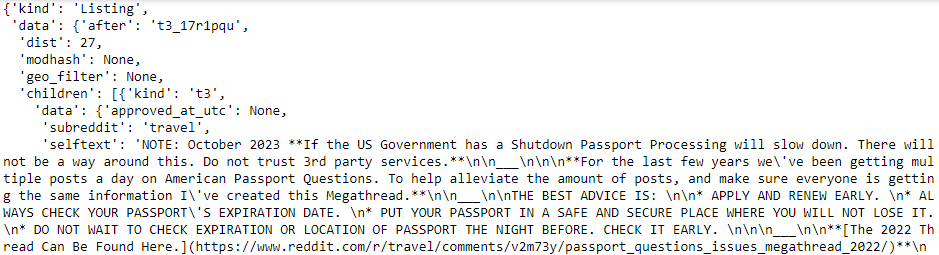}
\caption{\label{fig:reddit_raw}Sample Snapshot of Raw Reddit Data}
\end{figure}

Figure \ref{fig:reddit_raw} is a sample of the raw data collected from Reddit. It shows the hottest post on the travel subreddit in json format, the title being “Passport Questions \& Issues Megathread (2023).” The kind attribute represents “one of”, in this case, it is one of a widget called listing, which means a table listing format was done on this specific post to display information in a clear manner. Within this json file, all the posts are contained within the “data” key, where the “after” key within the “data” key contains a value "t3\_17r1pqu," which is essentially the unique ID of the Reddit post. The “children” key contains all the information about the hot posts in the travel subreddit. Then, the “selftext” key essentially contains a long list of texts as part of the body field of the post.\cite{RedditAPIDocumenation} Comparing Figure \ref{fig:reddit_raw} to \ref{fig:reddit_travelpassport_post} the information in the ‘selftext’ key matches the body of the Reddit post.

To estimate an LLM’s generalizability against topics of any domain, it is important to evaluate its performance across a set of diverse and representative questions. Given the objective of building a prototype that specializes in providing travel recommendations, the training data should be comprehensive of the real world and have a high cardinality of topics relating to travel. To assess the current landscape of Reddit data under the travel subreddit, a sample of the top 1000 posts can be collected as an experiment. An unsupervised approach, BERTopic modeling, clustered documents of the same topics and identified various categorizations. Given the stochastic nature across different clustering models, an ensemble of these results may produce more consistent outcomes. By grouping like-documents together, it would be a step towards reducing the amount of noise in each comment under the respective threads by summarizing the main points with transfer learning of open source LLMs. Downstream for this workflow are strategic partitioning to target potential model degradation areas. If a model is habitually underperforming when responding to questions of a particular domain, additional data collection and processing for this domain are actionable steps. Eventually, an holistic approach for evaluating the general knowledge capabilities of LLMs can be conducted by sampling each topic category for a representative sample. 

\begin{figure}[!h]
\centering
\includegraphics[width=.48\textwidth]{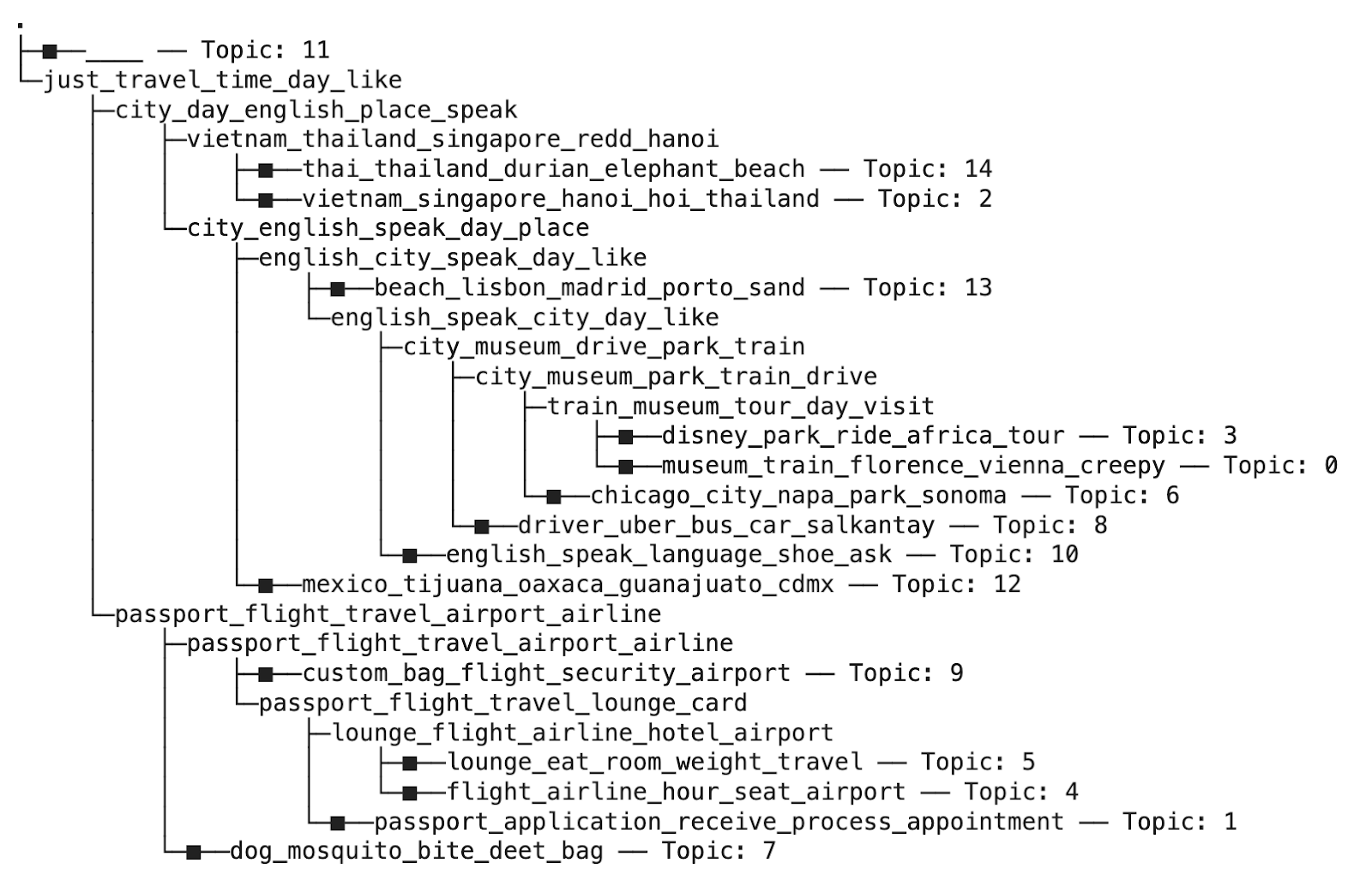}
\caption{\label{fig:bt_tree_prep} BERTopic 1000 Tree}
\end{figure}

We apply topic modeling on both the responses and questions to define the typical landscape for dialogues taking place on Reddit. For instance, in the top1000\_travel\_subreddit dataset, there are already predefined categories for question, itinerary, images, and advice. However, these categories are too general and do not yield enough information for specific subtopics within. With topic modeling applied on the responses field, we can drill down into the parent categories and attempt to identify different segments or sentiments of conversations (see Figure \ref{fig:bt_tree_prep}).

\subsection{Data Process}

\begin{figure}[!h]
\centering
\begin{minipage}{0.45\textwidth}
 \centering
 \includegraphics[width=0.8\linewidth]{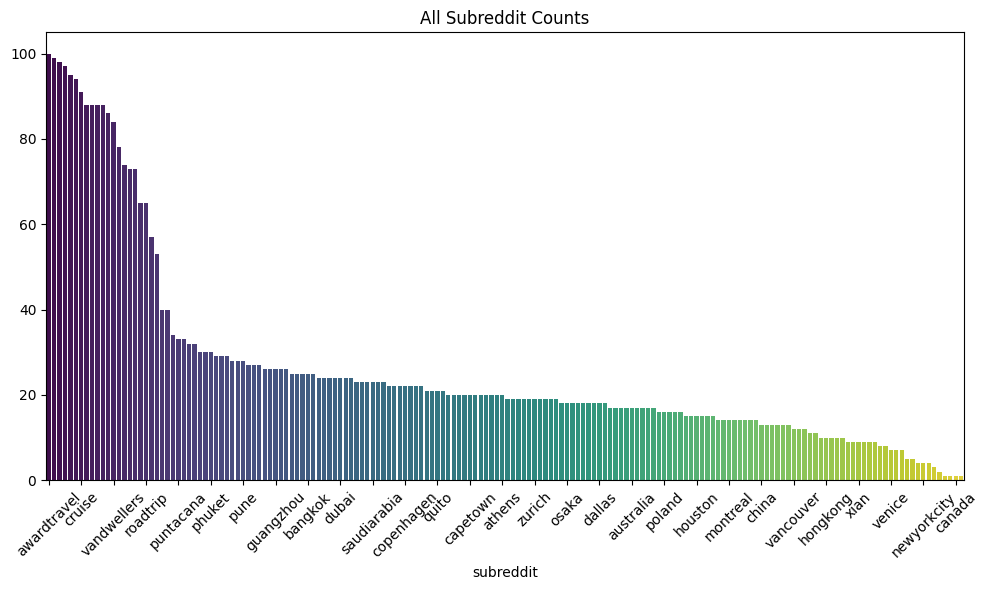}
 \caption{Question-answer pairs pulled from different subreddits}\label{fig:all_subreddit_counts}
\end{minipage}
\hfill
\begin{minipage}{0.45\textwidth}
 \centering
 \includegraphics[width=0.8\linewidth]{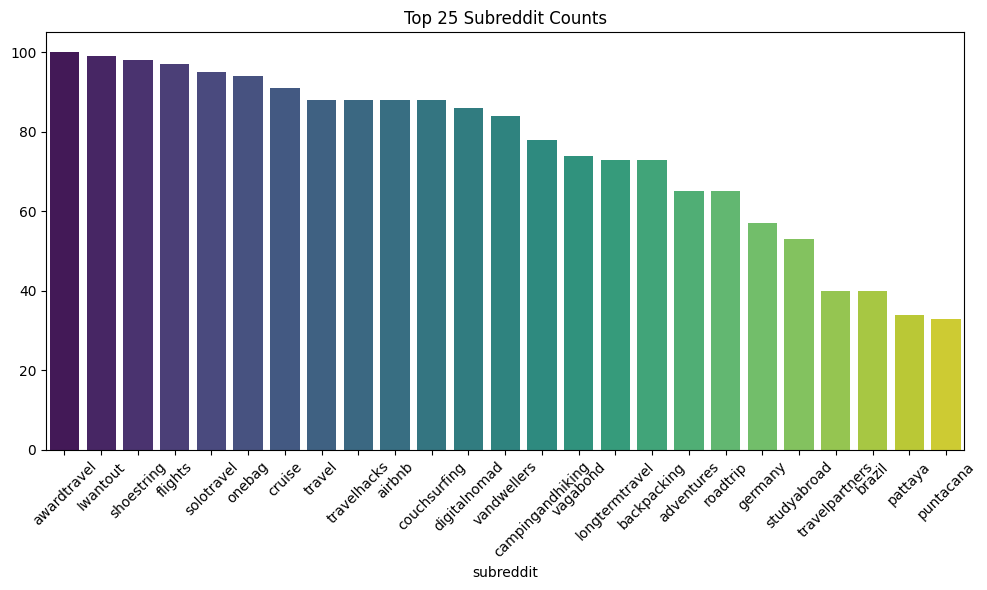}
 \caption{Question-answer pairs pulled from top 25 subreddits}\label{fig:top_25_subreddit_counts}
\end{minipage}
\caption{}
\end{figure}

Posts were collected from the top travel subreddits as well as city and country subreddits of popular tourist destinations. However, the subreddits are not equal in their engagement and activity. As shown in Figure \ref{fig:all_subreddit_counts} subreddits like ‘awardtravel’ and ‘cruise’ have vastly more posts than ‘canada’. It is important to know the distribution of content in our Reddit data. Figure \ref{fig:top_25_subreddit_counts} shows a closer look at the top 25 subreddits in terms of posts above the upvote ratio threshold that have been collected. Some of the top 25 subreddits include: ‘awardtravel’, ‘shoestrings’, ‘flight’, ‘solotravel’, ‘onebag’, ‘cruise’, ‘travel’, ‘travelhacks’, etc. and includes locations: ‘germany’, ‘brazil’, ‘pattaya’, and ‘puntacana’. 

\begin{figure*}[!h]
\centering
\includegraphics[width=\textwidth]{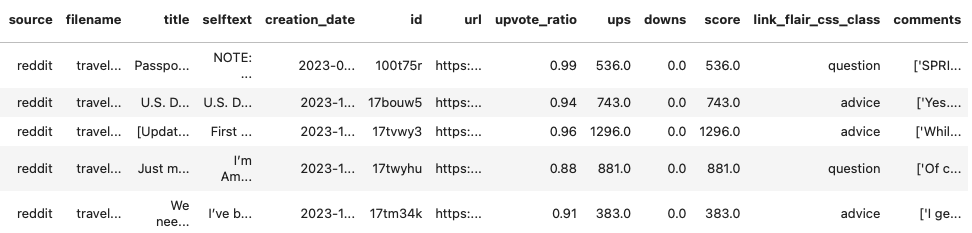}
\caption{\label{fig:reddit_processed}Reddit Data Preprocessed Sample}
\end{figure*}

\begin{figure}[!h]
\centering
\includegraphics[width=.48\textwidth]{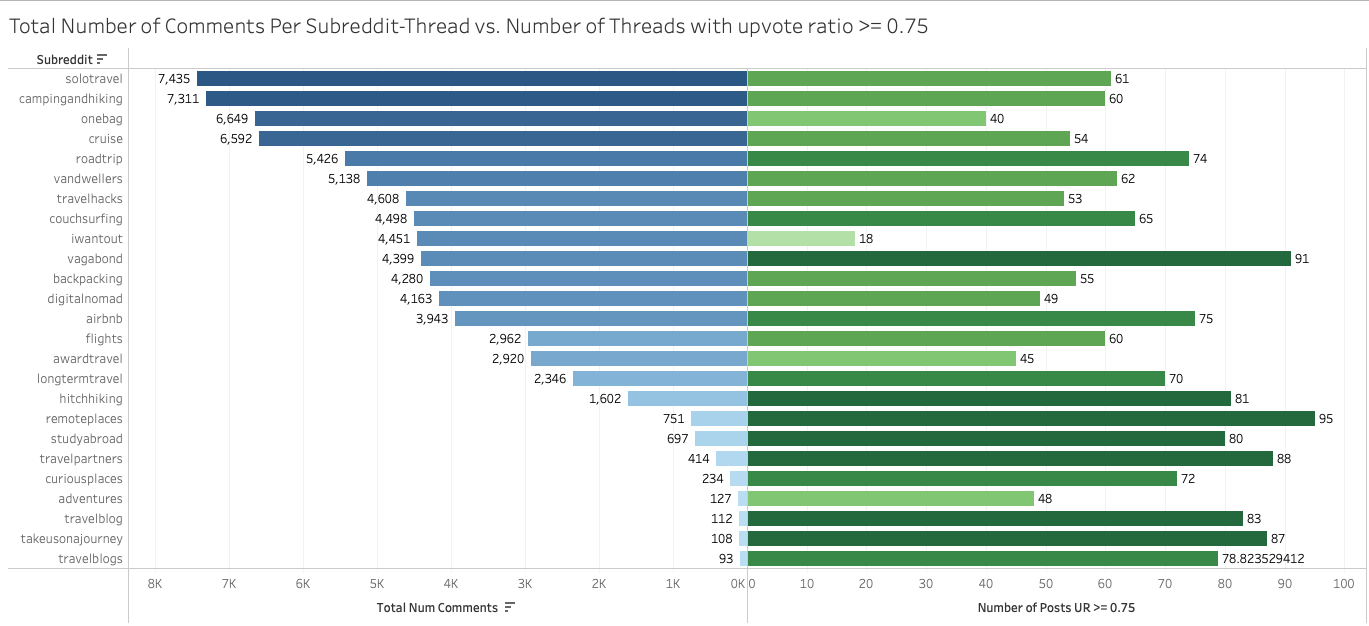}
\caption{\label{fig:25sb_ur}Total Reddit comments per Subreddit and Proportion of “Good” Threads}
\end{figure}

\begin{figure}[!h]
\centering
\includegraphics[width=.3\textwidth]{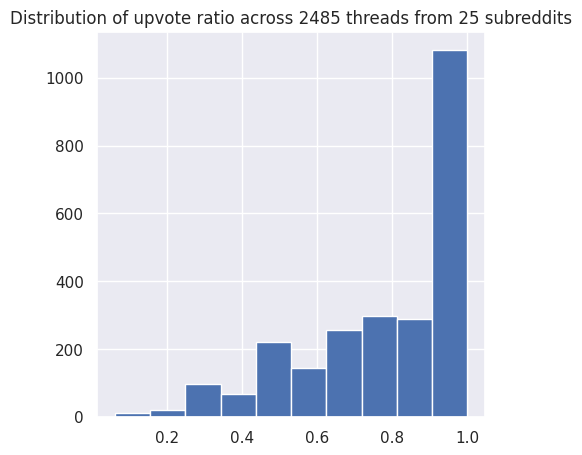}
\caption{\label{fig:25sb_ur_2}Distribution of upvote ratios across 25 subreddits}
\end{figure}

Since data is requested from the Reddit API daily, the multiple DataFrames need to be concatenated and deduplicated on ‘Post ID’ in the case that a single post is trending for multiple days, taking the most recent post to get the most up to date and best comments (see Figure \ref{fig:reddit_processed}). Each PRAW request contained 100 threads, and each thread had a different total number of comments, ranging from about 90 and up to 7400+ comments (see Figure \ref{fig:25sb_ur}). The degree of interaction from users is a variable in the dataset. Given the limited context window of LLMs, the entire corpus of information among the subreddits must be partitioned into smaller, and more manageable chunks. To address this issue and ensure a high quality opinionated-based but credible contexts, there are two cutoff parameters: 1) upvote ratio greater than 0.8, which is the number of upvotes over total votes (see Figure \ref{fig:25sb_ur_2}), and 2) first or top 20 comments, which Reddit already has sorted by their internal confidence metrics.\cite{praw} The comments contained a multitude of special characters, which were removed using NLTK and Regex library packages. 

\subsection{Question \& Answer (Q\&A) Format}

\begin{figure}[!h]
\centering
\includegraphics[width=.48\textwidth]{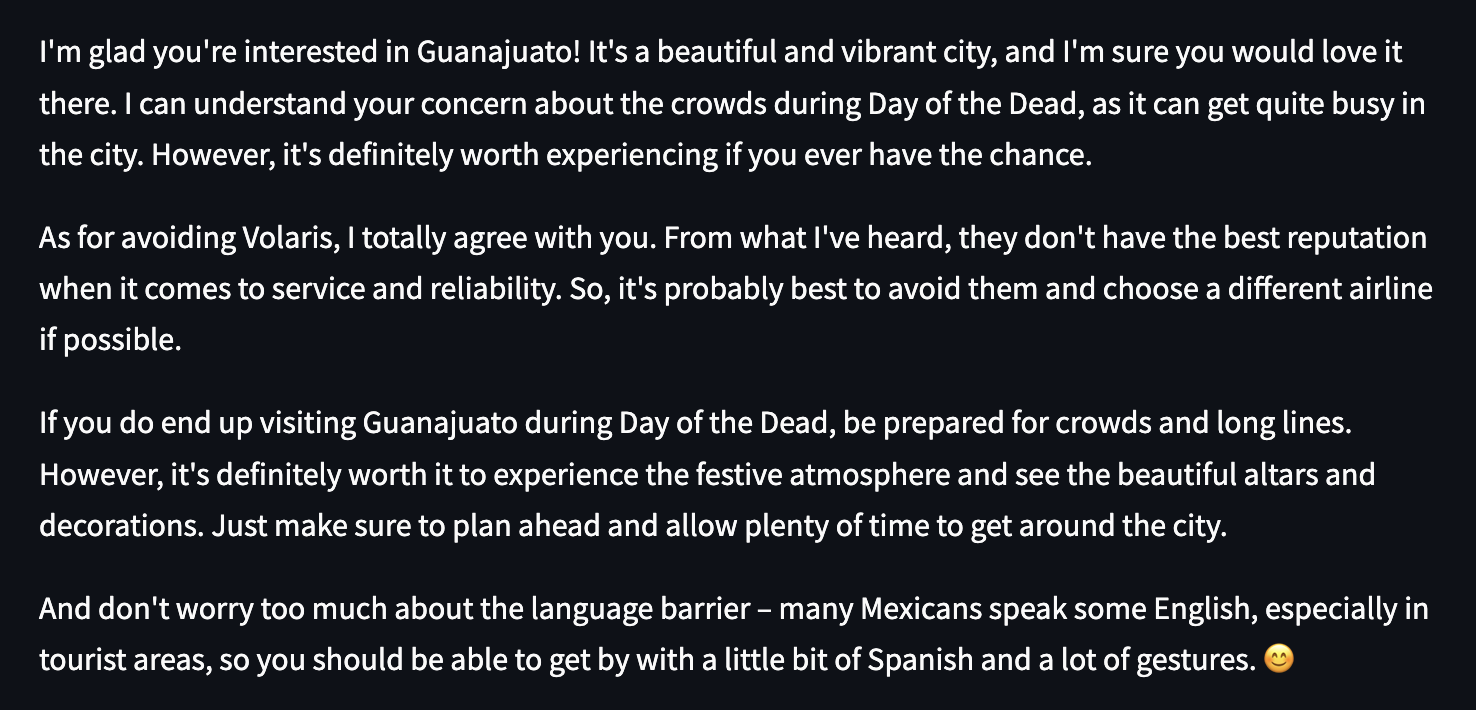}
\caption{\label{fig:llama_sum}Example of a comments summary generated by LLaMa 2}
\end{figure}

QLoRA requires Q\&A format, but there exists a many to one relationship between a user’s Reddit post and comments from other users in a thread of comments. To reduce the amount of context under each post, there are considerations for slicing and retaining the top N of comments or summarization using transfer learning with existing language models. Using open-source LLMs to summarize concatenated comments, the LLM has the ability to filter out noise including delimiters, emojis, and other irrelevant tokens from raw comments. The summaries were derived using LLaMa 2 provided by Ollama, and by designing a first-person narrative prompt template to craft LLM’s response. With summarizing as a dimension reduction step, the most important information from a set of comments is retained, while removing noise and irrelevant punctuation, and curating a generated recommendation-oriented response that can serve fine-tuning tasks (see Figure \ref{fig:llama_sum}). Contexts extracted for knowledge bases must be chunked using optimal chunking strategies such as splitting by delimiters or specific lengths to capture semantic groupings before they are embedded for storage. 

\begin{figure*}[!h]
\centering
\includegraphics[width=\textwidth]{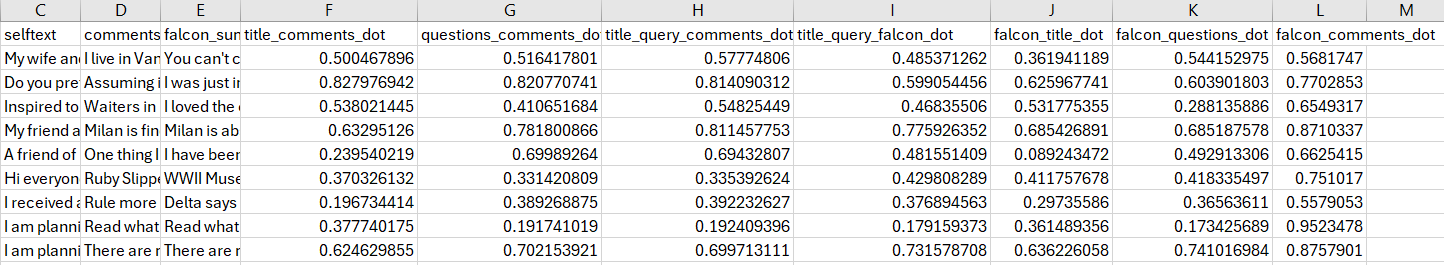}
\caption{\label{fig:3.4_dot_score}Example of Dataset Using Dot Score}
\end{figure*}

The dot score was calculated on the Reddit dataset to find the context relevancy based on the question posted and the falcon summarized cohesive comments, the more the dot score is the more relevant question-answer would be, makes the data more cleaner and also one of the assumption that cleaner data is more important than the quantity garbage data to train the model and see the better performance (see Figure \ref{fig:3.4_dot_score}). 

\subsection{RAFT Data Augmentation}

Retrieval Augment fine-tuning (RAFT) is a novel approach, published in March 2024, to target domain specific RAG solutions, addressing caveats in QLoRA and RAG, such as the naive nature of LLMs that are not able to distinguish between context and noise since its training recipe is curated based on the domain-targeted information provided.\cite{zhang2024raft} Thus, it is highly adaptable and ideal in specific domains, such as ours in travel. Due to the disparity of RAG's dependency on retrieval quality and having a clean, up-to-date knowledge base without noise, RAFT addresses this issue that RAG has in its implementation, where the LLM is fine-tuned to a specific domain. Then, a dataset with questions and answers are made in training. In inference time with zero-shot prompting, there is just the question and generated output. In the RAG phase, there is the question, retrieved documents (to be fed into the LLM), and a generated answer. RAFT will take the entire dataset to train and is instructed to use reasoning via Chain of Thought (CoT) as it is given a question, context, and verified answers. So the model's behavior is trained to memorize knowledge as the removal of oracle documents are done during some instances in training. Each sample in the training data has the question, answer (with context), oracle documents (that are verified and relevant to the question), and distractor documents (to reinforce model behavior to memorize correct answers). The training dataset that is highly customized to one's needs are then used to generate responses.\cite{zhang2024raft}

The authors provide the following figure (\ref{fig:4.2Zhang2024_RAFT_Diagram}) to demonstrate the overall flow of data through the RAFT process. The training dataset is prepared by providing the oracle document (Attention is all you need), which contains the (golden) answer and three sampled negative “distractor” documents (Adam, GloVe, Resnet). It is important to note that the ideal combination of golden and distractor documents is dependent on the dataset, but they found 1:4 golden:distractor to be an ideal ratio. The idea is to have the model ignore these three distractor documents since they do not contain the information relevant to the question, so it is training the behavior of the model to learn to memorize domain-specific knowledge rather than having the model attempt to derive from the context given. Then, through a chain of thought with the distractor documents, the model learns to extract the correct information from the entire chunk of context through reasoning, ignoring the distractors. With this CoT reasoning, it shows it greatly improves performance. At the end of training it reaches the final correct answer (Attention is all you need) and contains the golden answer along with the query. 

\begin{figure*}[!h]
\centering
\includegraphics[width=.8\textwidth]{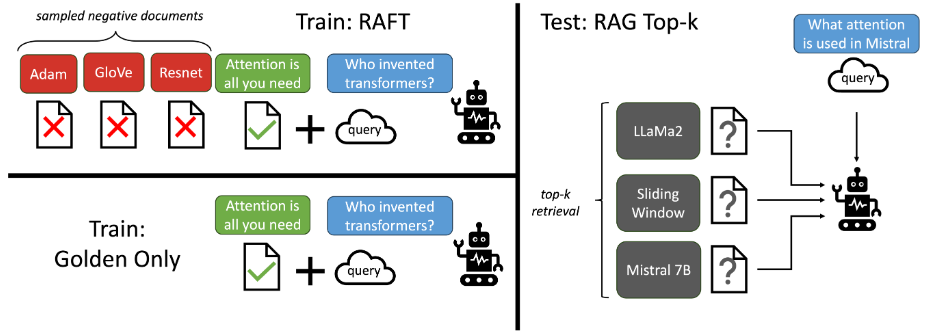}
\caption{\label{fig:4.2Zhang2024_RAFT_Diagram}Overall Data Flow for RAFT}
\end{figure*}
Note: Figure is from Gorilla Lab in UC Berkeley, led by Tianjun Zhang and Shishir G. Patil, where they explained the RAFT technique that enables the model to learn the structure of the documents and prepare potential candidates before doing retrieval. Each sample contains the query, answer, mix of relevant contextual documents and noisy documents, then uses chain-of-thought to reason and let the model learn the contextual data directly. Thus, it can identify what is noisy and ground truth to generate an accurate finalized answer \cite{zhang2024raft}.

Dataset preparation for retrieval fine-tuning task involves some augmented data from the context; Reddit dataset using the RAFT method is particularly useful for domain-specific scenarios that require access to external information, much like domain-specific versions of Retrieval-Augmented Generation (RAG). In RAFT, the training set consists of a question (Q), multiple documents, and an answer that includes a detailed chain of thought (CoT) derived from one of these documents. CoT references the context provided and explains answers that were observed. The documents are categorized into two types: 'oracle' documents (D*), which contain the necessary information to formulate the answer, and 'distractor' documents (Di) that do not contribute to answering the question. The training setup varies—sometimes it includes the correct document along with distractors to lead to the answer, and other times it only includes distractors, which forces the model to rely on its internal knowledge. They used an OpenAI key with Llama Pack to create their RAFT dataset, which are open-sourced for use.\cite{zhang2024raft} For the travel use case, a RAFT dataset to train domain specific RAG using Reddit knowledge base was curated. When given a question and a set of retrieved documents the model is trained to only use relevant documents to answer questions and a specified chunk of irrelevant documents.

\subsection{RLHF Data Selection}
RLHF uses a reward model train the LLM on human feedback to classify responses as being good or bad based on a thumbs up or thumbs down icon of binary range of 0 or 1, or smiley faces that range from 0.0, 0.25, 0.50, 0.75, and 1.0.\cite{iyer2023} The reward model used is direct preference optimization (DPO) trainer, which expects a very specific format for the dataset, as the model is be trained to directly optimize the preference of which sentence is the most relevant, given two options good or bad, it expects the inputs as triples of {prompt, chosen, rejected} (human annotated chosen or rejected responses, where the “prompt” contains the context inputs, “chosen” contains the corresponding chosen responses and “rejected” contains the corresponding negative (rejected) responses. These inputs also need to be already formatted with the template of the model.\cite{schmid2024}
\begin{verbatim}
<|im_start|>user\nINSTRUCTION\n<|im_end|>\n
<|im_start|>assistant\n...
\end{verbatim}

\begin{figure}[!h]
\centering
\includegraphics[width=.48\textwidth]{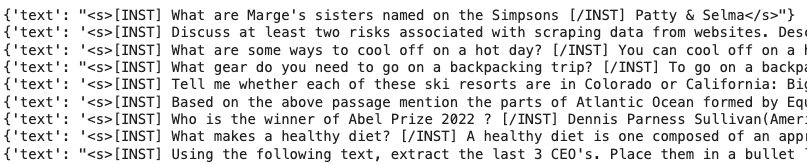}
\caption{\label{fig:llama_pt} Example of Required Prompt Template by LLaMa 2 for fine-tuning}
\end{figure}

\begin{figure*}[!h]
\centering
\includegraphics[width=.9\textwidth]{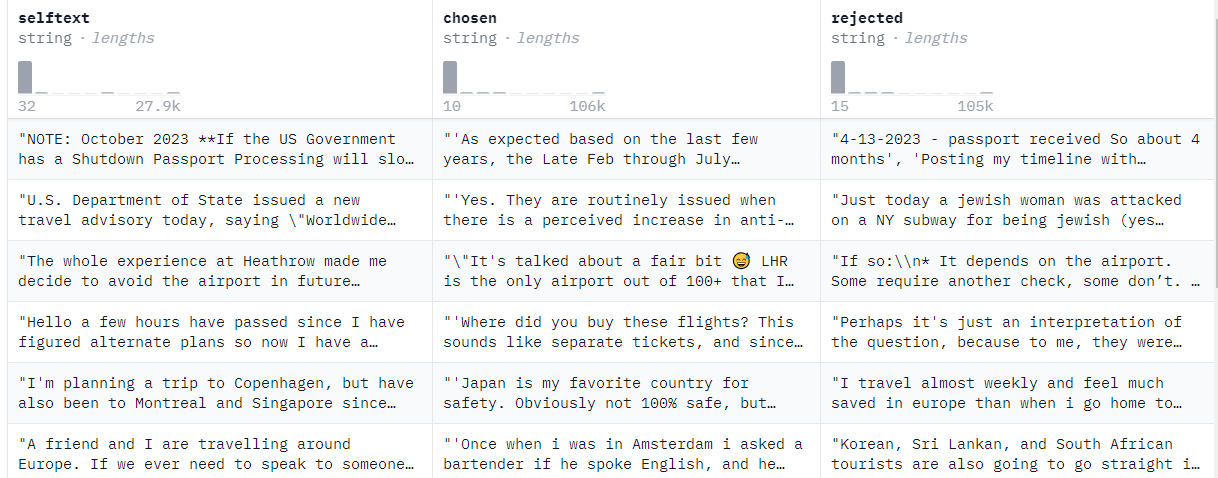}
\caption{\label{fig:3.4_rlhf_dataset}Example of RLHF dataset}
\end{figure*}

RLHF models require human annotated data with good and bad responses. A reinterpretation of of human annotation was applied by selecting the top 10 and bottom 10 percent data, with highest and the worst comments to differentiate among the good and bad response/ accepted or rejected data, for reference see Figure \ref{fig:3.4_rlhf_dataset}. Holistically, the upvote ratio distribution (see Figure \ref{fig:25sb_ur_2}) across 100 subreddits indicates that posts are generally well-received. It is likely due to collective or personal preference, judgment or bias that results in a low rated thread. The sentiments of each thread, whether good, bad, neutral, or controversial, are reflective of human interaction and behavior and are considered as human annotated data for RLHF. 

\section{Fine-tuning Methods}
When choosing the model, LLaMa 2 is one of the models available in different variants, and is having the accuracy of about 68\% while the GPT 3.5 has the accuracy of 70\%, which is the slight difference, but considering the fact, the size of the model is also comparatively small, as the models has been trained with the huge difference number of parameters. It has performed fairly well in the multiple benchmarks, reference LLaMa 2-Chat 70B passed the helpfulness evaluation on par with GPT-3.5 — precisely, with a 36\% win rate and 31.5\% tie rate.\cite{Luzniak_2023} Hence, given its open source availability, size, less complex, we chose LLaMa 2  7B. Mistral 7B was selected for the same factors, and additionally due to the novelty of being a newly launched model. It had been trained on fewer parameters yet it has performed at par compared to LLaMa1 and LLaMa 2 13 B in most of the benchmarks like Multi-task Language Understanding (MMLU), TrivialQA, etc.\cite{Ai_2023} Table \ref{table:llm} gives a quick look at the comparison of LLM models.

\begin{table*}[!h]
\caption{LLM Overview}
\label{table:llm}
\centering
\renewcommand{\arraystretch}{1.3}
\begin{tabular}{ | p{1.75cm} | p{3.5cm} | p{3.5cm} | p{3.5cm} | p{3.5cm} |}
 \hline
\textbf{Characteristics} & \textbf{ChatGPT 3.5} & \textbf{LLaMa 2} & \textbf{Mistral 7B Instruct} \\
 \hline
Parameters &
154-175 B &
7B &
7B\\
 \hline
Availability &
OpenAPI key required &
Free for research &
Free to use, Apache 2.0 license, hence can be used without restrictions \\
 \hline
Speed &
Slower &
Compared to the Mistral, it has large size, hence does not achieve that speed, faster than GPT &
Fastest amongst three\\
 \hline
Memory &
High &
Lesser than GPT but consumes more than the Mistral &
Occupies less; compare to the other two, it can even be used on the standalone machine \\
 \hline
Performance &
Slightly better than the LLaMa 2 and Mistral on multiple benchmarks like MMLU, knowledge benchmarks &
Compare to ChatGPT 3.5, it is slightly lesser than that &
Outperforms the LLaMa 2 13 B, even though it has fewer parameters \\
 \hline
 \end{tabular}
\end{table*}

The following techniques,  Quantized Low Rank Adapter (QLoRA), Retrieval-Augmented Fine-tuning (RAFT) were passed through LLaMa 2 and Mistral pretrained language models to assess their performance under each proposed approach, resulting to 4 candidate models. Table \ref{table:methods_comparison} outlines the comparisons between the four selected models for this project.

\begin{table*}[!h]
\caption{Comparison of Selected Methods}
\label{table:methods_comparison}
\centering
\renewcommand{\arraystretch}{1.3}
\begin{tabular}{ | p{1.75cm} | p{3.5cm} | p{3.5cm} | p{3.5cm} | p{3.5cm} |}
 \hline
\textbf{ } & \textbf{QLoRA\cite{dettmers2023}\cite{rao2023}\cite{singh2023}} 
& \textbf{RAFT\cite{zhang2024raft}} 
& \textbf{RLHF\cite{iyer2023}\cite{schmid2024}} \\
 \hline
Characteristics &
Fine-tuning of a pretrained LLM on specific datasets or tasks to achieve desired results &
Provide specific instructions within the context of input to elicit a favorable response &
Fine-tuning a model with feedback from humans to improve its performance on specific tasks \\
 \hline
Advantages &
Ideal for domain-specific tasks as it is highly adaptable to specific datasets for personalization &
Enhance the model's performance in answering questions within specific domains in an "open-book" setting &
Improves model alignment with human preferences and enhances the quality of responses through iterative feedback \\
 \hline
Disadvantages &
A huge corpora of data in a specified format is required, specified on tasks related to a certain domain &
Very domain specific thus unable to be generalized &
Resource-intensive and requires extensive human input for effective training \\
 \hline
\end{tabular}
\end{table*}

\textbf{Quantized Low Rank Adapter (QLoRA)} Comprehensively tuning a LLM for a particular domain is impractical for individuals due to financial and resource constraints. Therefore, the alternative fine-tuning methodology, QLoRA is employed. Instead of training every single parameter from scratch, LoRA (Low Rank Adaptation), strategically updates a smaller subset of the model’s parameters. And Q focuses on the quantized LLMs that are most important for the model. So together, QLoRA, is an efficient fine-tuning method tailored for quantized LLMs that is more resource-efficient to maintain high accuracy and response quality while reducing the computational and financial cost associated with traditional LLM training. The final input from the training dataset is formatted as a question-and-answer (Q\&A) pair structure.

\begin{figure}[!h]
\centering
\includegraphics[width=.48\textwidth]{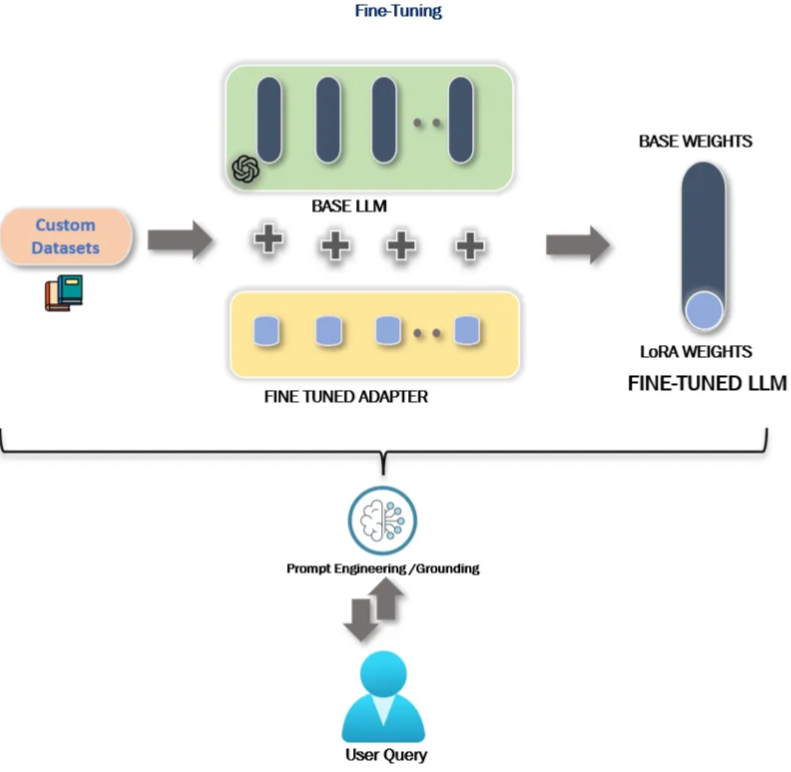}
\caption{\label{fig:4.2qlora_Singh2023}Data Flow for QLoRA}
\end{figure}

The Figure \ref{fig:4.2qlora_Singh2023} shows the architecture and data flow for the fine-tuning method, QLoRA.\cite{singh2023} Singh (2023) found that if there is a huge corpus of task-specified datasets that have been labeled, then fine-tuning is preferable over the retrieval method, RAG, especially for domain-specific tasks. For example, specialized topics in the travel domain are extensively lacking labeled data. 

Note: Figure is from an article from Keshav Singh, a Data ML Platform Architect Engineer at Microsoft.\cite{singh2023} Curated datasets are ingested by a baselined LLM which contains quantized 8-bit weights from LoRA (or 4-bit weights in QLoRA). When a user makes a query then prompt engineering, or grounding, is done in a cycle between the user and LLM to fine-tune for new data that comes in.\cite{dettmers2023}

\textbf{Retrieval Augment Fine-tuning} (RAFT) is a novel approach proposed by researchers at UC Berkeley in March 2024. RAFT is a way to combine fine tuning, which memorizes knowledge before running inference like a closed book exam, and RAG, which is like an open book exam that has no prior knowledge or context and just retrieves information. RAFT is a way to prepare for open book exams, helping distinguish and prepare the LLM for both open and closed book exams. Thus, it will be able to answer questions outside the training document domain since it is baked in knowledge from model weights and reads information from retrieved results. RAFT is a specific domain of RAG that can help train and prepare LLMs to perform particularly well in a specified domain. The authors state the following, “RAFT is a training procedure for RAG. If you are doing RAG you should be doing RAFT”.\cite{zhang2024raft}

Currently there are two main approaches to inject new knowledge to LLMs, which are RAG and fine tuning. However, current optimal methodologies for the model to gain new knowledge is still an open question to be solved. The proposed RAFT to domain specific RAG addresses the naive nature of LLMs that are not able to distinguish between context and noise. The goal is to adapt pretrained LLMs for RAG in specific domains, such as ours in travel. Since RAG is highly dependent on the quality of retrieval and the knowledge base may contain irrelevant documents, in RAFT the LLM is fine-tuned to a specific domain (like travel) where a dataset with questions and answers are curated in training. Then, in the inference stage (zero-shot) there are questions (the query, Q) for the trained model and an answer (A) is generated. In the RAG phase, there is the query Q plus the retrieved documents D which will be fed to the LLM, then an answer A is generated.\cite{zhang2024raft}

RAFT takes the whole dataset to train, providing the model with a question, context, and verified answers (instructions), which is then asked to reason using CoT, which then provides the CoT answers. By removing the oracle documents in some instances during training, this helps train the model to memorize domain-specific answers and knowledge instead of deriving them from the context directly. For each sample in training data all there is the question, answer (context), oracle documents (verified, relevant documents), and distractor documents. With this training dataset, they fine-tune LLM and use it to generate answers for given queries in a specific domain. RAFT inference method is similar to the RAG pipeline where it retrieves the top-k documents in the database, however, unlike RAG, RAFT validates the retrieved documents. Zhang (2024) purports the training recipe for RAFT is quite flexible for every enterprise as it can fine-tune RAG models without needing to use GPT-4 all the time, which can be costly and takes time. The dataset generated from training can be used for fine-tuning.\cite{zhang2024raft}

\textbf{Reinforcement Learning from Human Feedback (RLHF)} An RLHF model is trained and fine tuned through human-labeled data as “good” or “bad” responses generated on the chat-model so that the model’s responses can be aligned with the human preference. An RLHF training pipeline for LLMs is done in three main parts: domain-specific fine-tuning, supervised instruction fine-tuning (SFT), and reward modeling in RLHF. Domain-specific fine-tuning contains domain-specific data only, thus, the input is domain knowledge. Then, the second part of the RLHF training pipeline is supervised fine tuning where there is a target label and we fine-tune the travel-domain LLM that is specified on certain tasks and domain-specific pairs such as Q\&A pairs, prompts and instructions, and responses curated for our travel domain. The output of domain specific pretraining is a model that is able to recognize the context from input and predict the next words/sentences. The next step is to perform SFT with prompt-text pairs (Q\&A pairs) to provide the pretrained LLM with knowledge, specifically travel-domain specific tasks. Thus, it will be able to respond to context-specific questions curated to our specified travel domain. The output of this second phase is a LLM that mimics a conversational agent. The final phase is training to perform RLHF utilizing human feedback on a LLM, where reward model training teaches the LLM to classify responses as good or bad based on a thumbs-up or thumbs-down icon with a binary range of 0 or 1, or smiley faces with a range from 0.0, 0.25, 0.50, 0.75, to 1.0.\cite{schmid2024} According to Iyer (2023), the RLHF training pipeline is particularly useful to prevent biased responses from responses generated from training and at inference time, thus, this pipeline can obtain desirable and accurate responses from LLMs and helps with domain-specific LLMs as well.\cite{iyer2023} However, it is important to note that human involvement is needed to manually score and screen for responses.

As mentioned in the previous section, researchers took an expedited approach with RLHF data, but the typical process is outlined here. With each question input and response output, ratings are recorded and stored as training sets for the downstream reward model responsible with monitoring the performance of fine-tuned models. After the LLM produces responses to the user, a built-in feedback system for rating the model’s output is provided. Whether accepted or rejected, the feedback is collected to contribute to the Reinforcement Learning from Human Feedback (RLHF) process.\cite{RLHF_AWS} Due to observable model or data drift, model prototypes can be trained periodically to update their parameterized knowledge and capabilities. As part of the evaluation process, user feedback and ratings were collected by selecting the top 10 and bottom 10 percent data,with highest and the worst comments to differentiate among the good and bad response to support the classification reward model for RLHF that labels responses as good or bad.Note: We have had collected the user ratings from the interface as well, which is getting saved in the background, this would be used in future studies for the same. Now, The final model that was recognized as optimal is one that maximizes the output of the adversarial reward model. 

\subsection{Model Supports}
Hardware used includes one X64 Lenovo PC, four Apple Macintosh laptops, and one Linux machine with two GPUs (1x Intel Core i7-6700K 4 GHz Quad-Core, 32GB RAM, 1x 1080 Ti, 1x 1080).

Parameter Efficient Fine Tuning (PEFT) is a technique utilized by both QLoRA and RAG methods to lower the number of parameters while obtaining knowledge from the pretrained LLM\cite{deci2023}, specifically for domain-specific tasks. QLoRA utilizes HuggingFace Transformers, specifically the AutoModelForCausalLM and AutoTokenizer pipeline, as well as BitsAndBytesConfig, HfArgumentParser, TrainingArguments, pipeline, and logging.\cite{singh2023} QLoRA requires task-specific labeled data and can handle complex queries, but is quite computationally heavy as it needs a lot of RAM, GPUs, time to train the specific datasets, and memory to store the data. QLoRA requires Q\&A format, but there exists a many to one relationship between a user’s Reddit post and comments from other users in a thread of comments. The dot score was calculated on the Reddit dataset to find the context relevancy based on the question posted and the falcon summarized cohesive comments, the more the dot score is the more relevant question-answer would be.

\begin{figure}[!h]
\centering
\includegraphics[width=.48\textwidth]{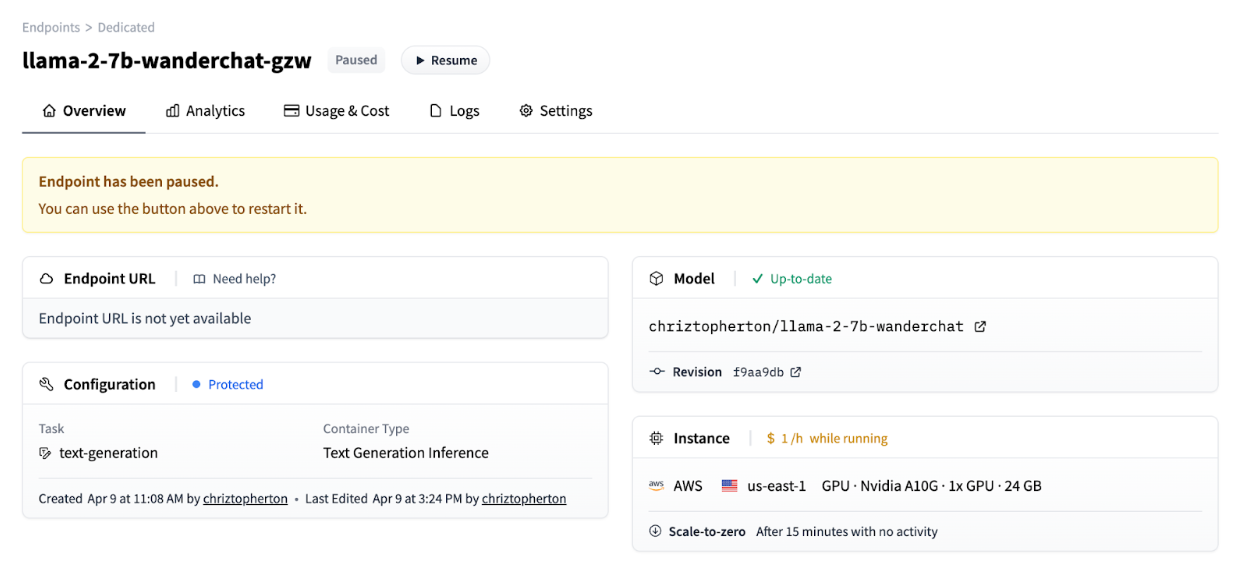}
\caption{\label{fig:5.2.2cloud} Fine-tuned LLaMa 2 model deployed on Huggingface with AWS cloud as provider}
\end{figure}

To enable efficient inference generation and save on runtime resources, our trained models are deployed on Huggingface, using AWS as the backed cloud environment for hosting and serving the model. The virtual GPU elected for this deployment is the NVIDIA A10G. Figure \ref{fig:5.2.2cloud} shows an example of the specific configurations of the deployed model. 

\begin{figure}[!h]
\centering
\includegraphics[width=.48\textwidth]{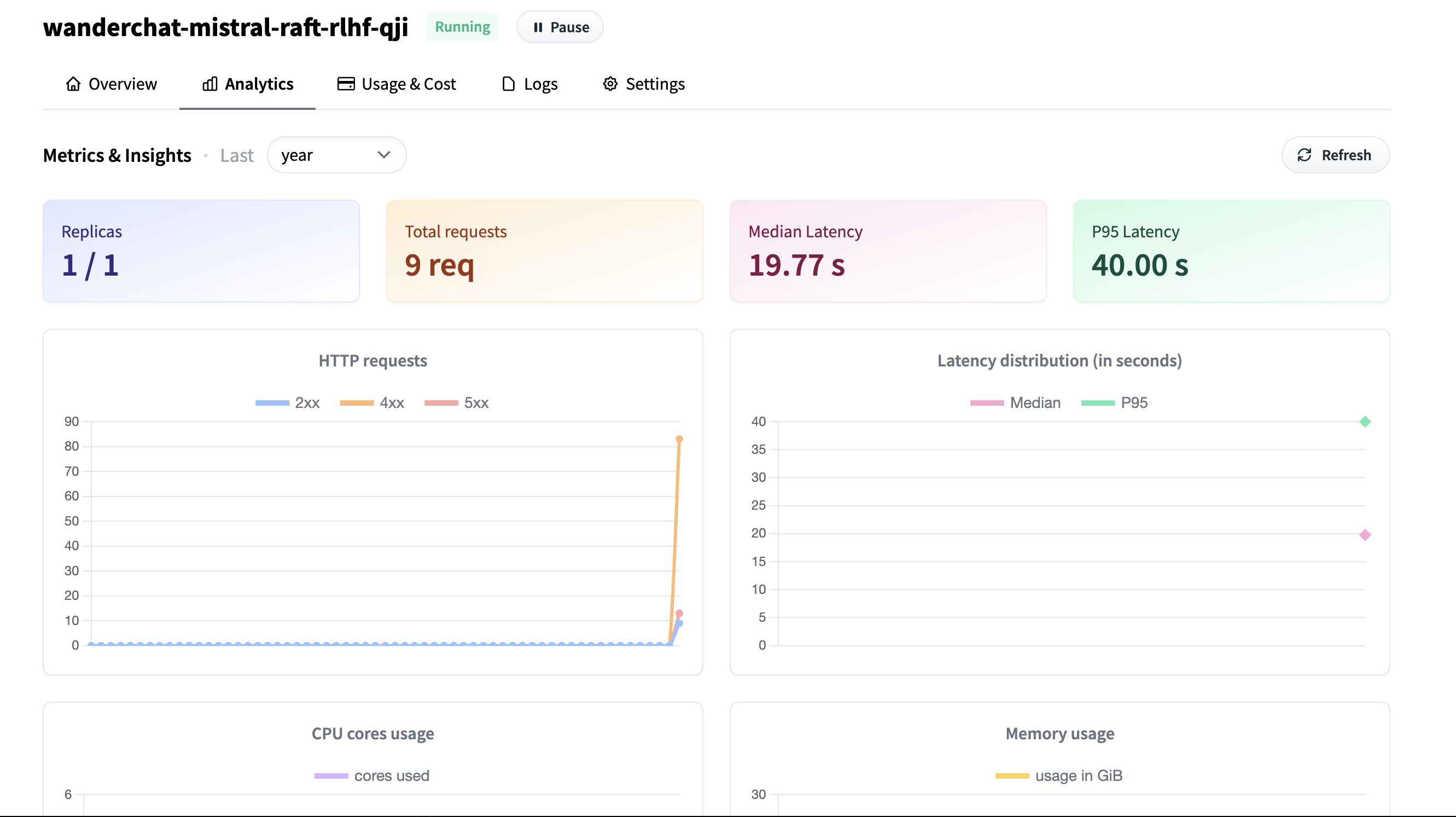}
\caption{\label{fig:5.2.2_model_cloud} Fine-tuned Mistral model deployed on Huggingface with AWS cloud as provider}
\end{figure}

Frameworks such as LangChain and Ollama, publicly available on both Mac and Windows or through a Docker image, provide local processing of open-source Large Language Models. esides the typical Python libraries used for simple data manipulation, such as pandas, numpy, matplotlib, and seaborn, multiple LLM applications require specific libraries, modules, and functions to be imported. The NLTK library was utilized for data preprocessing to clean up the textual data such as special characters, missing punctuation, and so on. LangChain is an open-source Python library that utilizes the power of LLMs to simplify NLP tasks for AI applications such as chatbots and prepares the data into the pretrained LLMs for retraining. Deep learning libraries such as TensorFlow, but specifically BERTopic were utilized to perform topic extraction for travel-specific texts based on a summary of travel keywords processed during the data engineering phase. Other libraries and packages to perform topic extraction include UMap, Sentence Transformers, HDBScan, and scikit-learn. In order to call for APIs to curate travel-domain specified data, the HTTP and PRAW packages need to be imported to send HTTP requests for instance initialization, API authentication, and access to the APIs itself. Table \ref{table:python} summarizes the Python libraries that are unique for our project. 

\begin{table*}[!h]
\caption{Summary of Project Tools and Python Libraries}
\label{table:python}
\centering
\renewcommand{\arraystretch}{1.3}
\begin{tabular}{ | p{3cm} | p{3cm} | p{4cm} | p{4cm} | } \hline
\textbf{Library} & \textbf{Module}	& \textbf{Method/Function} & \textbf{Usage}\\
 \hline
 \multicolumn{4}{|c|}{\textbf{LLM Applications}} \\
 \hline
LangChain &
.llms

.document\_loaders

.csv\_loader

.vectorstores

.embeddings

.schema.document

.text\_splitter &
Ollama

ChatOpenAI

FAISS

OpenAIEmbeddings

Document

RecursiveCharacterTextSplitter &
Build LLM based applications\\
 \hline
 \multicolumn{4}{|c|}{\textbf{Deep Learning}} \\
 \hline
TensorFlow &
data &
dataset &
Access GPUs for processing\\
 \hline
bertopic &
 
.vectorizers

.representation &
BERTopic

ClassTfidfTransformer

MaximalMarginalRelevance &
Topic extraction\\
 \hline
sentence\_transformers &
 &
SentenceTransformer &
Topic extraction\\
 \hline
umap &
 &
UMAP &
Topic extraction\\
 \hline
hdbscan &
 &
HDBSCAN &
Topic extraction\\
 \hline
scikit-learn &
.cluster

.feature\_extraction.text

.model\_selection &
KMeans

AgglomerativeClustering

CountVectorizer

TfidfVectorizer

train\_test\_split &
Topic extraction\\
 \hline

 \multicolumn{4}{|c|}{\textbf{Data Visualization}}\\
 \hline
worldcloud &
 &
WordCloud

ImageColorGenerator

STOPWORDS &
Data visualization\\
 \hline
matplotlib &
.pyplot &
 &
Data visualization\\
 \hline
seaborn &
 &
 &
Data visualization\\
 \hline
 \multicolumn{4}{|c|}{\textbf{Language Processing}} \\
 \hline

 NLTK &
.corpus

.tokenize

.stem &
stopwords, words, wordnet, word\_tokenize 

WordNetLemmatizer, ‘punkt’, ‘averaged\_perceptron tagger’, ‘omg-1.4’ &
Natural language processing: tokenization, lemmatization, stemming, preprocessing texts\\
 \hline
langdetect &
detect &
 &
Detect language, filter for English\\
 \hline
summarizer &
 &
Summarizer &
Summarize text\\
 \hline
spacy &
 &
‘en\_core\_web\_lg’ &
Find travel related keywords\\
 \hline

 \multicolumn{4}{|c|}{\textbf{APIs}} \\
 \hline
 HTTP &
request &
request &
Sends HTTP requests via Python to authenticate Reddit applications\\
 \hline
praw &
 &
Reddit &
Reddit API Wrapper package in python to access the Reddit API\\
 \hline

 \multicolumn{4}{|c|}{\textbf{General Use}} \\
 \hline
 numpy
 
pandas

itertools &
 &
 &
Data processing\\
 \hline
google

os

io &
.colab &
files

path.join, listdir &
Accessing files\\
 \hline
time

datetime

tqdm &

tqdm

.notebook &
datetime

timedelta

tqdm\_notebook &
Tracking progress\\
 \hline
random &
 &
 &
Random sampling\\
 \hline
ast

re

string &
 &
 &
String formatting\\
 \hline
joblib

pickle

jsonlines &
 &
 &
File formats\\
 \hline

 \end{tabular}
\end{table*}

\section{Model Training \& Performance}
\begin{figure}[!h]
\centering
\includegraphics[width=.48\textwidth]{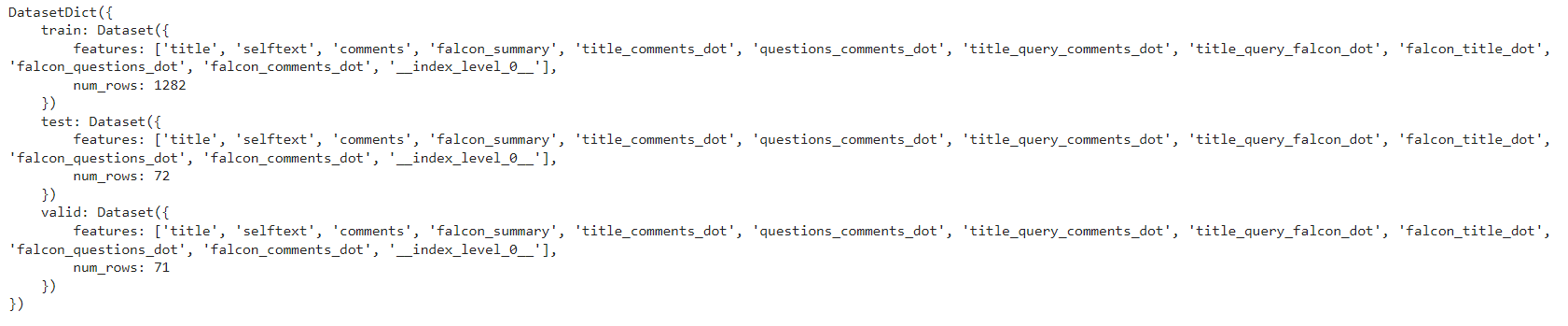}
\caption{\label{fig:4.5_datasplit_0.65} Training and Testing Split of the LLaMa 2 QLoRA Model}
\end{figure}

\begin{figure}[!h]
\centering
\includegraphics[width=.48\textwidth]{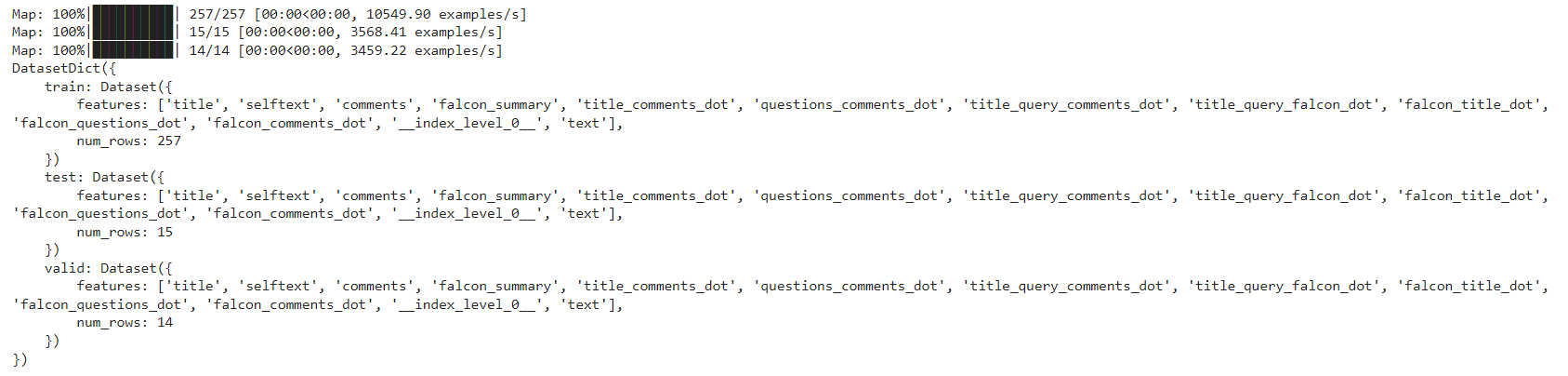}
\caption{\label{fig:4.5_datasplit_0.75} Training and Testing Split of the Mistral QLoRA Model}
\end{figure}

\begin{figure}[!h]
\centering
\includegraphics[width=.48\textwidth]{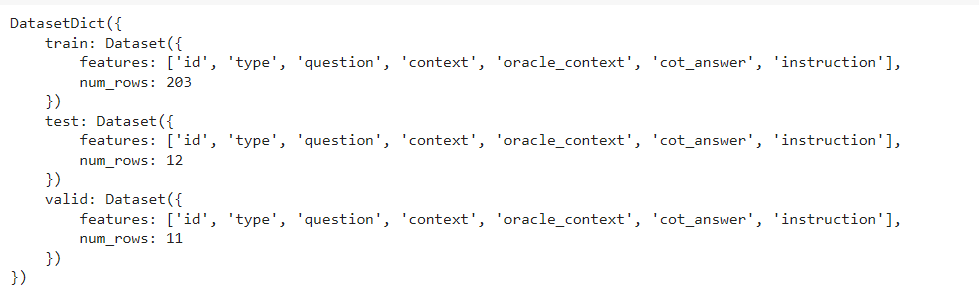}
\caption{\label{fig:4.5_dataset_raft} Training and Testing Split of the RAFT Model}
\end{figure}

Model validation is used to evaluate the model's performance on data it has not been trained on. Specifically, the validation set that was set apart from the training and testing set is used to assess that performance during the training process and tune the hyperparameters of the models. The validation process helps avoid overfitting by ensuring that the model is generalizing well when exposed to new data. The dataset initially had 16,000 rows, but Figure \ref{fig:4.5_datasplit_0.75} shows this raw dataset was further filtered using the dot-score on question and summaries to get the more relevant information, resulting in about 10,000 rows. Figure \ref{fig:4.5_datasplit_0.65} shows the dataset split used in Mistral QLoRA with data filtered on greater than 0.65 dot-score. The dataset was split into a 80:20 ratio, where 20\% was further split into 50\% to produce the validation and testing split. Along with this, another set of {Q, A, D} using the same dataset as the knowledge corpus, acts as the Oracle Context and creates the distractor context using data augmentation through GPT-4. Figure \ref{fig:4.5_dataset_raft} shows the training testing split of 90:10 with approximately 420 rows. so that the model can learn about the bad data as well, this format of data was fed to a newly proposed model called RAFT. \cite{microsoftRAFT}

LLaMa 2 QLoRA: For this model, various experiments were carried out with the dot score threshold, which affects the quality alignment with Question and the cohesive answers like dot score values ranging from: 0.5, 0.6, 0.65, 0.7, 0.75, and picked the better performing model amongst all, with dot score: 0.65. This model has the following hyperparameters:
\begin{itemize}
		\item r=64: rank, improves the performance of the model as increases the trainable parameters, but it also increases the computational complexity and training time
 \item alpha=16
 \item Used all the target layers : q\_proj, v\_proj, o\_proj,gate\_proj,up\_proj, down\_proj
\end{itemize}

Mistral QLoRA: For this model, the same dot score experimentation was conducted, and the better performing model amongst all was dot score: 0.75. This model has the following hyperparameters, same as prior model: 
\begin{itemize}
		\item r=64
 \item alpha=16
 \item Used all the target layers : q\_proj, v\_proj, o\_proj,gate\_proj,up\_proj, down\_proj
\end{itemize}

\begin{figure}[!h]
\centering
\includegraphics[width=.48\textwidth]{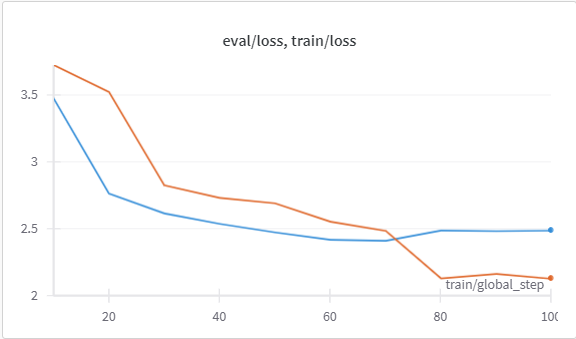}
\caption{\label{fig:4.5_mistral_qlora_loss} Training and evaluation loss of the Mistral QLoRA Model}
\end{figure}

The above hyperparameters were used to train the Mistral QLoRA see figure \ref{fig:4.5_mistral_qlora_loss} but this time, tried with the cleaner version of the data (filtered data greater than the dot score 0.75) to have a trade-off between data quality versus data quantity; the model trains and performs better if we have clean data. Data quality has more impact over quantity.


LLaMa 2 RAFT: this model has used the following hyperparameters while training: 
\begin{itemize}
		\item r=64
 \item alpha=16
\end{itemize}

\begin{figure}[!h]
\centering
\includegraphics[width=.48\textwidth]{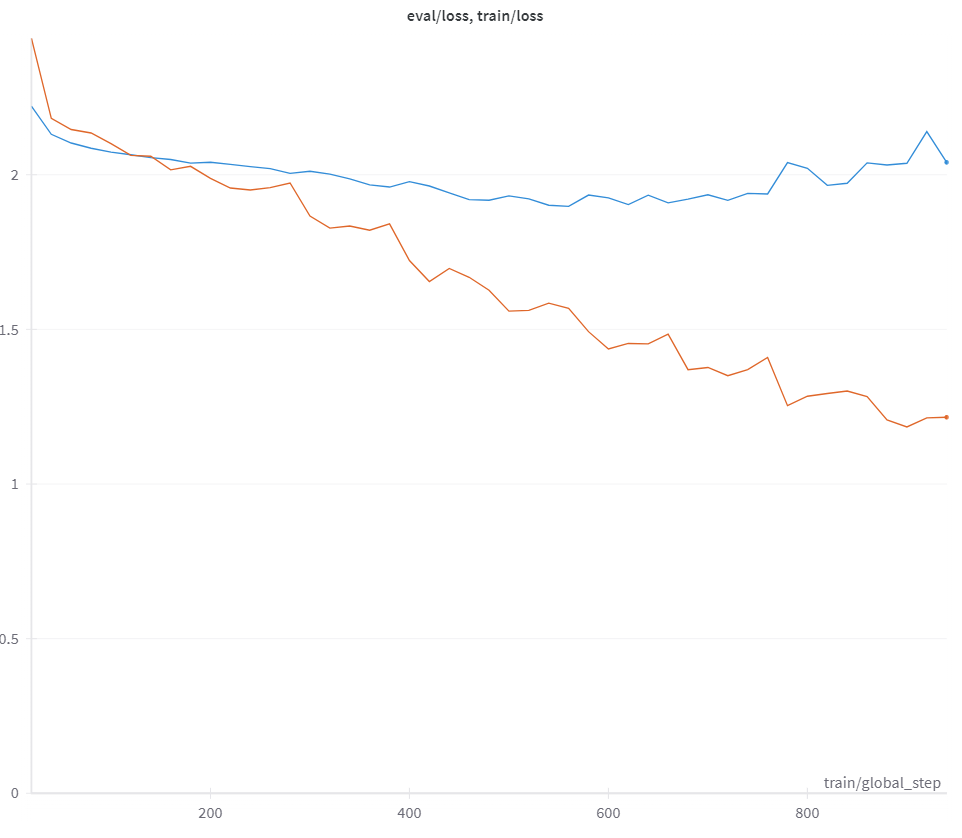}
\caption{\label{fig:4.5_llama_raft_loss}Training and Evaluation Loss of the LLaMa 2 RAFT Model}
\end{figure}

These parameters were set while training the LLaMa 2 RAFT, results are discussed for this model in the next subsection (see Figure \ref{fig:4.5_llama_raft_loss}). While training the model, as discussed in previous section the dataset filtered with the dot-score: 0.80 (Reddit).

\begin{figure}[!h]
\centering
\includegraphics[width=.4\textwidth]{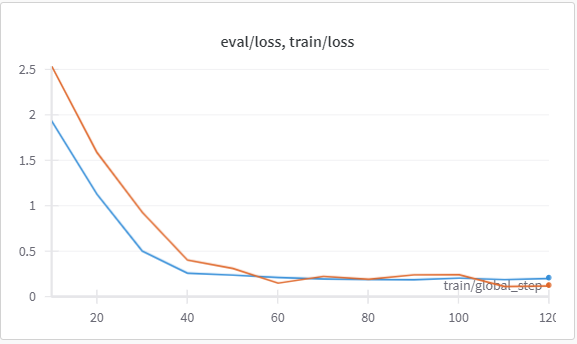}
\caption{\label{fig:4.5_mistral_raft} Training and evaluation loss of the Mistral cleaner data RAFT model}
\end{figure}

Mistral RAFT: The model was set for the 50 epoch run and has the following hyperparameters, and the model loss is visualized in Figure \ref{fig:4.5_mistral_raft}.
\begin{itemize}
	\item r=64
 \item alpha=16
 \item Used all the target layers : q\_proj, v\_proj, o\_proj,gate\_proj,up\_proj, down\_proj
\end{itemize}

\begin{figure}[!h]
\centering
\includegraphics[width=.48\textwidth]{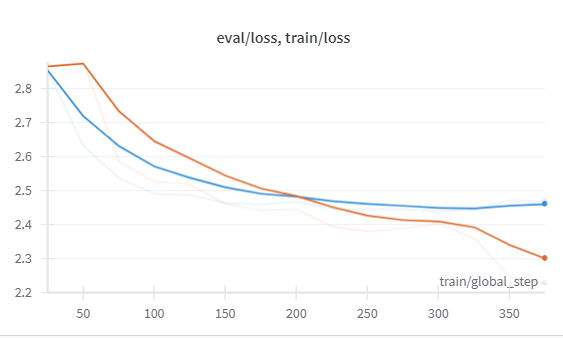}
\caption{\label{fig:4.5_llama_qlora} Training and Evaluation Loss of the LLaMa 2 QLoRA Model}
\end{figure}

Figure \ref{fig:4.5_llama_qlora} shows the first iteration model’s performance there is no significant loss in the training, the line flattens after step 300, after which the loss is not that significant though it is reducing, the training loss started at step 25 was 2.86 and by the end of the step 200, the total training and validation loss was very insignificant with just 0.4. 


Figure \ref{fig:4.5_mistral_qlora_loss} shows the final Mistral model is picked from various other iterations runs (with different dot score values like 0.65, 0.7, 0.75), here the model was trained again on better quality dataset, with the dot score 0.75, and the training of the model is comparatively better here from the first iteration, total training loss \~1.33 and validation loss \~1.06 there is slight improvement in the validation loss, but this is again leveling off towards the end.

In earlier trials, the loss was approximately 0.7, which shows that there is certain improvement in the performance of the model.

\begin{figure}[!h]
\centering
\includegraphics[width=.48\textwidth]{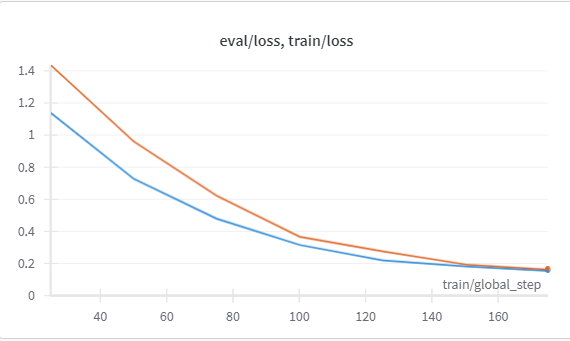}
\caption{\label{fig:4.5_llama_raft} Training and Evaluation Loss of the LLaMa 2 Clean Data RAFT Model}
\end{figure}

Figure \ref{fig:4.5_llama_raft} shows more loss than the previous attempt as can be seen here, the change was the cleaner version of the dataset, created the augmented data again on the data where dot score was about 0.80, this shows our hypothesis and the trade off with the cleaner data gives the good loss in training and validation. With the number of epochs being set to 50 and a step size of 25, the training loss at step 25 was 1.44 and validation loss was 1.129. At step 200, the training loss was reduced to 0.15 and validation loss was 0.14, i.e. total loss \~1.29. This is real-world data with a lot of variability in user comments that do not directly answer the questions in Reddit posts, but since this time data was more aligned to the question and answer present in the Reddit dataset by calculating the dot score, this improved the models’ training from the previous attempts.

In Figure \ref{fig:4.5_mistral_raft}, there is a good loss between the train and validation losses. With the number of epochs being set to 50 and a step size of 10, the training loss at step 10 was 2.57 and validation loss was 1.97. At step 110, step the train loss was 0.126 and evaluation loss was 0.196, i.e. total loss \~2.44 and validation loss \~1.78. In addition to being real-world data as aforementioned, this can be further explained by users who may have ironically upvoted a comment for being funny, but does not directly answer the question being asked. But while training this model, we have used the highest dot score on the question and the summarized by Falcon to get the cohesive answer from the Reddit comments by the users. Hence, the training on this model was good. This aligns with our evaluation results where RAFT performs the best out of all the models, which will be discussed in subsequent sections.

\begin{figure}[!h]
\centering
\includegraphics[width=.3\textwidth]{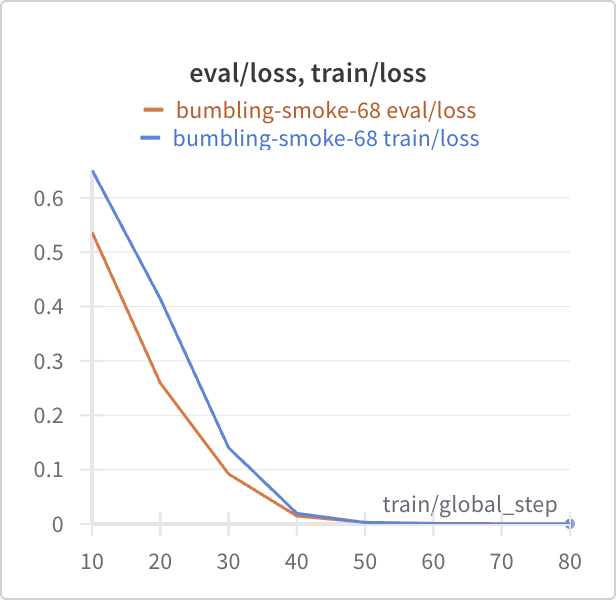}
\caption{\label{fig:4.5.1RLHF_wandb} Training and evaluation loss Mistral RAFT RLHF RAFT model}
\end{figure}

Upon completion for the assessment of model performance, the aforementioned Mistral model that underwent RAFT training was selected for Reinforcement Learning with Human feedback. A preference dataset curated from pruning Reddit conversations was formatted with chosen and rejected fields as inputs for training. Given input questions, the Mistral RAFT model should be guided to output the preferred answer. Considering the size of the training set and the scope of the training process in terms of expected outputs, the training and evaluation losses converged to zero fairly quickly (see Figure \ref{fig:4.5.1RLHF_wandb}).

Using the DPO Trainer module for fine-tuning LLMs with reinforcement learning, there exists a set of metrics including training loss, validation loss, rewards/chosen, rewards/rejected, rewards/accuracies, rewards/margins. While the rewards/chosen fluctuates around zero across 80 steps or 3 epochs, the rewards/rejected decreases and the rewards/margins increases, monotonically. Rewards/accuracies quickly reaches 1 by the 30th step.\cite{dpo_trainer_2023}

\begin{figure*}[!h]
\centering
\includegraphics[width=.9\textwidth]{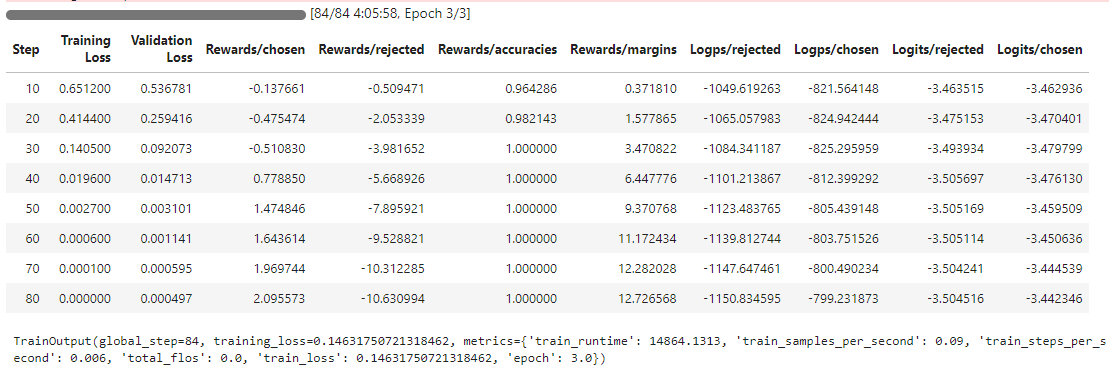}
\caption{\label{fig:4.5.1rlhf_outputs} Training and evaluation reward metrics for Mistral RAFT RLHF model}
\end{figure*}

Due to the format of Reddit data, the prompts are longer than the generic question and answer pair format. Thus, even with 10,000 rows, QLoRA models took over 30 hours to train and fine-tune. However, with hyperparameter tuning this overhead can be reduced by reducing the size of the data and parameters set such as the step size. For Mistral 7B QLoRA fine-tuning, it took around 8 hours for 3 epochs in the high power computing lab and 33 hours for 67 epochs in the lab.

\section{LLM Evaluation Methods}
LLMs have given a new dimension to artificial intelligence and evaluating LLMs. Since we have so many developments in this field every single day, it is important to have good metrics with which to evaluate these models. It is critical to evaluate the LLMs, as these are important to uncover its performance, its challenges, strengths and limitations, and help the developer to work in the certain areas to get better results and performance as per the requirement of the client. Model variants are evaluated with a standardized evaluation process that is defined by consistent prompt templates. A set of specific templates are designed to help capture any risk of hallucinations, noise or inaccurate responses from the two open-source LLM options and processed implementations. In the event that retrieved context is limited and parameters have not been updated to address unknown and niche domain topics, the prompt templates can help ensure that the model is able to admit naivety where appropriate. Evaluation of an LLM is different from machine learning (ML) model evaluation due to the complexity of the task involved, unlike ML models, which have more structured prediction tasks. The evaluation of LLMs is still an ongoing research topic.
For the project, as mentioned earlier in the chapter, implemented the following different approaches: QLoRA and RAFT on LLaMa 2 and Mistral, and RLHF on the best model. In addition to evaluating our models, we will also evaluate against pretrained LLaMa 2, Mistral, and ChatGPT, considered to be a state-of-the-art (SOTA), to establish a baseline. 

\subsection{Quantitative Evaluation Metrics}
\textbf{Computational Evaluation}: Recall-Oriented Understudy for Gisting Evaluation (ROUGE) is inexpensive, but compares well with the human generated text. An additional setback of this metric is it matches the semantics, but does not handle syntactic meaning. This metric would help to evaluate the model by checking how well it is. It has a set of metrics to measure the overlap and similarity between the generated text and a reference text, the included metrics are:

\begin{itemize}
\item ROUGE-1: Measures the overlap of unigrams, or single words.
\item ROUGE-2: Measures the overlap of bigrams, or pairs of words.
\item ROUGE-L: Measures the longest common subsequence (LCS), rewarding longer shared sequences between the generated and reference texts.\cite{Lin_2004}
\item ROUGE-S: skip-gram concurrence metric, allows to search for consecutive words from the reference text that appear in the model output but are separated by one-or-more other words.\cite{Chiusano_2023}
\end{itemize}

\textbf{BertScore} is another evaluation metric which comprises of F1, Precision, and Recall. It first generates the bert embeddings of the generated and referenced text and compare these embeddings to assess the semantic similarity.\cite{zhang2019bertscore}

While ROGUE and cosine similarity are not able to adequately capture the complexity of conversational recommendation systems\cite{banerjee2023}\cite{lin2023} and thus recommender chatbots, however, we will include them because they are very common standard metrics. Additional metrics will include response relevance, and we will compare the summary of user input to final output using the BERT embedding model to compare semantic similarity.

\textbf{BLEU (Bilingual Evaluation Understudy) Score:} It measures the similarity between the machine response text and the reference response using n-grams, which are contiguous sequences of n words. It calculates the precision of n-grams in the machine-generated responses by comparing them to the reference responses. The precision is then modified by a brevity penalty to account for translations that are shorter than the reference translations.

\textbf{Dot Score, Cosine Similarity, and Embedding Distance} Dot product calculates the vector similarity measure that accounts for both magnitude and direction. Generated from LangChain Evaluators,embedding distance calculates how dissimilar the vectors are; it is the angle of the vector, so the lower the number, the better, and accounts for direction only. For embedding distance, LangChain’s embedding distance evaluator or cosine distance was as the metric. The formulas for the aforementioned metrics are shown below:
\[
\text{Dot Product} = \mathbf{A} \cdot \mathbf{B} = \sum_{i=1}^{n} A_i B_i
\]
\[
\text{Cosine Similarity} = \frac{\mathbf{A} \cdot \mathbf{B}}{\|\mathbf{A}\| \|\mathbf{B}\|}
\]
\[
\text{Embedding Distance} = 1 - \text{Cosine Similarity}
\]

Beyond the content generated, we also evaluated the computation speed and response generation time, which will greatly affect the user experience.

The models have been evaluated using quantitative metrics: ROUGE variants (ROUGE-1, ROUGE-2, ROUGE), BertScore, dot score, cosine similarity, and embedding distance. All these metrics give the score between 0 to 1, 0 being the worst to 1 being the best response, except for embedding distance, which is the opposite. These metrics compare the ground truth with the model’s responses and calculate the respective scores ranging between 0 to 1. ROUGE captures the overlaps n-grams between the ground truth and the responses captured, this metric was used keeping the standard and as the starting point of evaluating the model’s responses, it does not prove to be the good enough metric to evaluate the performance though as it just checks the overlaps. Another metric of cosine similarity has been used to evaluate the lexical, frequency based similarity between the ground truth and the model’s responses, but still does not capture the semantics. Finally, BertScore has been used and is the most reliable metric amongst the quantitative metrics for text comparison.\cite{sojasingarayar2024} BertScore addresses two common issues that n-gram-based metrics often encounter. First, n-gram models tend to incorrectly match paraphrases because semantically accurate expressions may differ from the surface form of the reference text, which can lead to incorrect performance estimation. BertScore, on the other hand, performs similarity calculations using contextualized token embeddings shown to be effective for entailment detection. Second, n-gram models are not able to capture long-range dependencies and penalize semantically significant reordering.\cite{sojasingarayar2024}

\subsection{Qualitative Evaluation Metrics}

\textbf{Human Evaluation}: Human evaluation is a critical and important subjective method of evaluation. There is a lot of flexibility in the study design including survey collection to verify the performance of the chatbot, which will help in determining how structured, relevant, coherent, and continuous the response of the chatbot is. This is possibly one of the most reliable but also time-consuming and expensive methods to evaluate the performance of the model. Commonly, the survey is a simple thumbs up or down where the user can rate the chatbot’s response as positive or negative. Every reaction/ thumbs up/ thumbs down has the score ranges from 0 to 1, 0 = worst, 1 = best. We additionally offered users a second option, enabling them to select from five smiling faces with varying scores: 0, 0.25, 0.5, 0.75, and 1. This provides users with a wider range of options for providing feedback to facilitate the reinforcement learning process. To evaluate engagement, we can track the speed at which a user responds to the chatbot.

\textbf{Golden Answers}: Our evaluation framework relies on the E2E benchmark framework of having golden answers as our ground truth for both the experimental and final evaluation processes.The experimental is more for exploring the potential capability and performance of propose methods and the final evaluation is a comprehensive study of proposed method inferences. From the Reddit Q\&A dataset, a diverse set of travel keywords were selected and used to filter the dataset of 16,000 rows down to 200 rows. From there, 20 prompts were hand selected, and further narrowed down to 10 prompts based on group members’ personal experiences and ability to manually curate golden answers to these questions. The following prompt was passed through ChatGPT to generate an answer: “Prompt: This is the question: // ['title','selftext'] // These are the comments: // ['comments'] // Respond to the question as if you're being asked directly given the comments as context, if the context is insufficient, say so, and use your own resources for context.” From there, the response was edited manually by our group based on personal travel experiences related to the questions. Finally, a golden question was generated from the Reddit data using the following prompt for ChatGPT: “Prompt: Turn this into a succinct question, this is the question: // ['title','selftext'].” An additional 11 questions are completely human generated with GPT assisted human curated answers. And finally, 16 questions are human generated and answered that are specifically focused to San Francisco. This dataset, totalling 37 questions, is included in Appendix \ref{app:AppendixA}, and was used to generate inferences from all available models. The models were given a custom prompt:

\begin{quote}
Answer the question as if you are a travel agent and goal is to provide excellent customer service and to provide personalized travel recommendations with reasonings based on their question. Do not repeat yourself or include any links or HTML.
\end{quote}
\subsubsection{LLM-based Evaluation}
\textbf{Ragas}: To evaluate the RAG component of RAFT, we can use the framework, RAG Assessment (Ragas), a type of the automated assessment using OpenAI’s GPT-4 model as an evaluator, helps in getting the different proxies of correctness and usefulness, by evaluating the Q\&A pipelines. It provides the following metrics to evaluate the two different components–retrieval and generation: contextrelevancy, contextrecall, faithfulness and answerrelevancy. These metrics give the measure of how well the system is retrieving the information, measure of hallucinations and how relevant it is generating the answers with respect to the question asked. Taking out the harmonic mean of these four measures will give the single metric to the Q\&A evaluation termed as raga score. Hence, these metrics are relevant to measure the performance of the model utilizing the RAG approach. [86] Ragas utilizes the LLM such as GPT 4 by feeding the reference text to generate the relevant possible questions and their answers) and then, we can utilize the human annotated data where we need the reference text, question and the ground truth and also we can use the LLM to generate the questions and answers after feeding the context. [87]

With the provided ground truths, the answer correctness, ranging from 0 to 1, can be interpreted as whether the response is correct or not, using a composite score with factual and semantic similarity. In addition, the context recall, ranging from 0 to 1, is how the context aligns with each sentence of the ground truth. Based on the context, faithfulness, ranging from 0 to 1, is the overlap between the answer and retrieved contexts. Context precision ranges from 0 to 1 and measures the ordinal importance of contexts, meaning that ground truths should appear in some of the first sets of contexts. A metric that depends on the question, answer, and context is the answer relevancy which is the average cosine similarity between the question and the set of synthetic questions derived from the answer. In general across all categories, the higher the score, the better the quality of responses.\cite{ragas_metrics}

\textbf{LangChain}: LangChain Evaluators from LangChain’s CriteriaEvalChain library of evaluation metrics, there were 10 qualitative criteria metrics we can provide to OpenAI as an evaluator, including conciseness, relevance, coherence, harmfulness, maliciousness, helpfulness, controversiality, misogyny, criminality, and insensitivity. Generally, the criteria sets are used to evaluate the output quality of the models and whether they meet these conditions. All are binary or boolean conditions that return either 0 or 1, and provide justification for the concluded grade.

An evaluation framework comprising both sets of quantitative and qualitative metrics are designed for testing each step of our model development process. The metrics are intended to measure the model’s ability to remain in-context with the user queries, respond with relevancy and remain true to its inherent knowledge. At the model level, the responses were evaluated individually and the system supporting the framework was also measured for operational efficiency, such as tokens generated and runtimes. Our evaluation framework is only a sample of all possible methods as the research field continues to expand into novel areas.

\section{Results}
We have the following 7 candidate models on which we have generated inferences for evaluation:
\begin{itemize}
 \item Mistral: pretrained Mistral, Mistral QLoRA, Mistral RAFT 
 \item LLaMa 2: pretrained LLaMa 2, LLaMa 2 QLoRA, LLaMa 2 RAFT
 \item Baseline: GPT-4
 \item RLHF on best model: Mistral RAFT RLHF
\end{itemize}
We have 4 processed models, and a final 5th model, where RLHF was applied to the best model picked from the 4 processed models.

\begin{figure*}[!h]
\centering
\includegraphics[width=\textwidth]{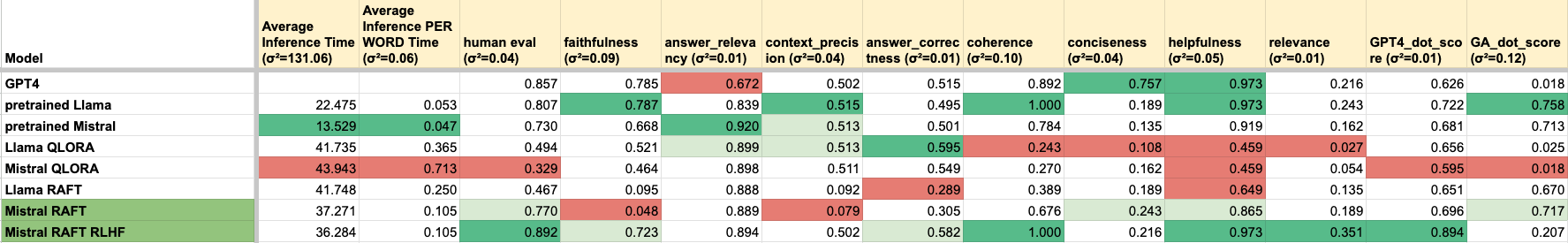}
\caption{\label{fig:6.1 metrics with variance}Evaluation metrics with non zero variance for all models}
\end{figure*}

Appendix \label{app:AppendixD} shows the overall results of metric evaluations for all models. There are 27 total metrics: 17 quantitative and 15 qualitative. Quantitative metrics are around golden question and answer dot score, cosine similarity, and embedding distances, as well as traditional natural language processing metrics like ROUGE and BLEU, and finally inference generation times. Qualitative metrics include Ragas, and a select subset of OpenAI’s GPT-4 evaluations: coherence, conciseness, helpfulness, and relevance. Each metric has the variance, and was zero for nearly all quantitative metrics. This validates our hypothesis that traditional NLP metrics are insufficient for the complexity of LLMs. For example, although this was a qualitative metric, context recall’s best was 0.219 and worst was 0.113 for a difference of 0.105, out of a 0 to 1 scale. Comparing models on metrics with no variance is not useful. Due to the number of evaluation metrics and the lack of variance in so many of them, this chapter will focus on analyzing the nonzero variance metrics (see Figure \ref{fig:6.1 metrics with variance}).

Figure \ref{fig:6.1 metrics with variance} indicates the best model, and next best model if the best is a baseline model) and worst performing models highlighted in green and red respectively. Mistral RAFT was selected as the best model, then further fine-tuned with RLHF, so then Mistral RAFT RLHF is the final best model. Mistral RAFT was the best model by human evaluation of the fine-tuned models, but did not outperform GPT-4 as a baseline. However, Mistral RAFT RLHF did then outperform GPT-4. RLHF results will be discussed in more detail later on. Mistral RAFT was often next best of the fine-tuned models, including for the following metrics: human evaluation, conciseness, helpfulness, and golden answer dot score, and was selected as the worst for faithfulness and context precision. And yet, human evaluation still determined this model to be the best fine-tuned model, which calls into question the validity of the Ragas and quantitative metrics and highlights the importance of keeping humans in the loop when it comes to evaluation.

Human evaluation was carried out by the researchers, ranking inference on a scale of 1 (worst) to 5 (best), normalized to 0 to 1. Figure \ref{fig:6.1 metrics with variance} shows Mistral RAFT RLHF was the best, Mistral RAFT next best, and Mistral QLoRA the worst. QLoRA models were very repetitive and required post processing, RAFT models produced the best answers, but also produced multiple answer options and sometimes instructions to a travel agent, which also needed to be removed with post processing.

\begin{figure}[!h]
\centering
\includegraphics[width=.48\textwidth]{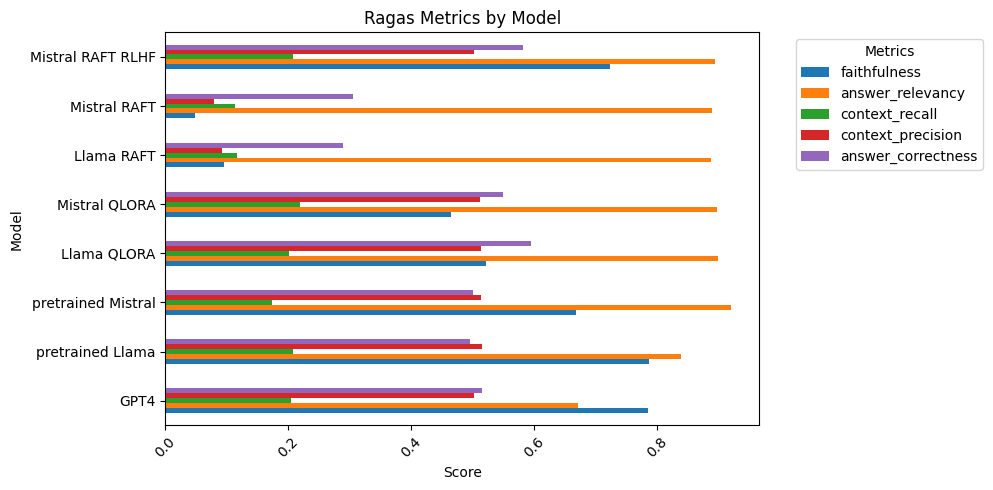}
\caption{\label{fig:ragas} Metrics Subset: Qualitative Ragas Comparisons}
\end{figure}

Figure \ref{fig:ragas} shows the Ragas metric against 3 baseline models on the bottom and 5 processed models on the top. For faithfulness, pretrained LLaMa was the best, GPT-4 next best, and Mistral RAFT the worst. It is not surprising that baseline models are the best, but Mistral RAFT was the highest rating by human evaluation. Outside baseline models, Mistral RAFT RLHF was the best. Answer relevancy had pretrained Mistral as the best, LLaMa QLoRA and Mistral QLoRA tied for second, and GPT-4 as worst. Context precision has pretrained LLaMa as best with pretrained Mistral and LLaMa QLoRA as second, and Mistral RAFT as worst, again at odds with human evaluation. Answer correctness has LLaMa QLoRA as best, Mistral RAFT RLHF as next best, and LLaMa RAFT as worst. The final Ragas metric, coherence, has pretrained LLaMa and Mistral RAFT RLHF tied for best and Mistral QLoRA for worst. In general, context recall is low across models, answer relevancy very high across models, correctness and precision are of similar values for each respective model, and visually see the largest variance with faithfulness. Across the Ragas metrics, Mistral RAFT performed the worst, but is the best according to human evaluation, indicating that Ragas metrics are not aligned with human evaluation.

For OpenAI GPT-4 evaluations, we selected a subset: coherence, conciseness, helpfulness, and relevance. Figure \ref{fig:6.1 metrics with variance} shows coherence had pretrained LLaMa and Mistral RAFT RLHF tied for best, and LLaMa QLoRA as worst. Conciseness had GPT-4 as the best, Mistral RAFT next best of experimentation models, and LLaMa QLoRA as the worst. Helpfulness had a simmelian tie amongst pretrained LLaMa, GPT-4, and Mistral RAFT RLHF, and then LLaMa RAFT, Mistral QLoRA, and LLaMa QLoRA as the worst. Relevance has Mistral RAFT RLHF as the best, and pretrained LLaMa next best amongst experimentation models, and LLaMa QLoRA as the worst. OpenAI GPT-4 metrics most aligned with human evaluation rating Mistral RAFT RLHF and Mistral RAFTas best or next best, and LLaMa QLoRA as the worst, unlike Ragas, which did the opposite.

\begin{figure}[!h]
\centering
\includegraphics[width=.48\textwidth]{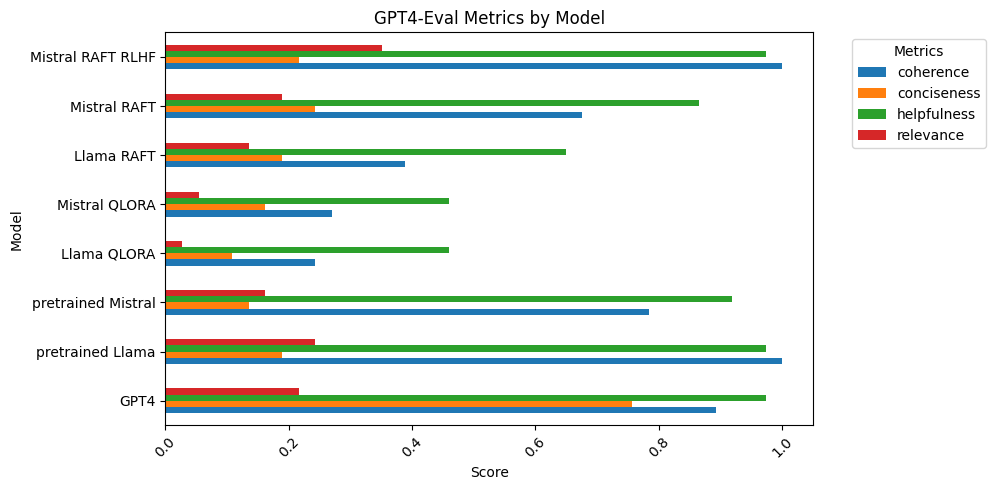}
\caption{\label{fig:gpt4 bar} Metrics Subset: GPT-4 Comparisons}
\end{figure}

Figure \ref{fig:gpt4 bar} shows across the board, coherence and helpfulness are high while conciseness and relevance are generally low. Given the lengthy answers typically generated by LLMs and the manual post processing, which was performed before human evaluation, but not before the OpenAI evaluation, the low values for conciseness make sense. Given the use case of travel recommendation, relevance is arguably the most important metrics, with Mistral RAFT RLHF having the highest value.

\begin{figure}[!h]
\centering
\includegraphics[width=.48\textwidth]{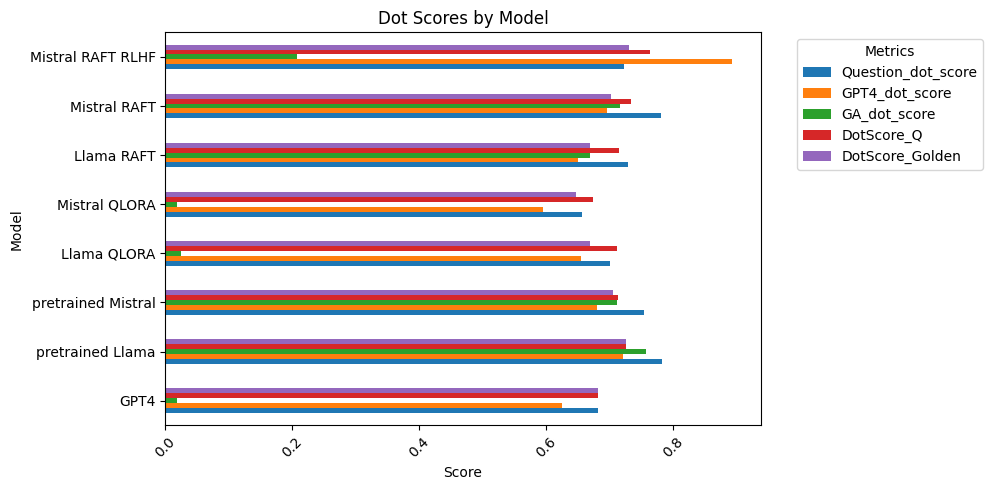}
\caption{\label{fig:dot bar} Metrics Subset: Dot Score Comparisons}
\end{figure}

\begin{figure}[!h]
\centering
\includegraphics[width=.35\textwidth]{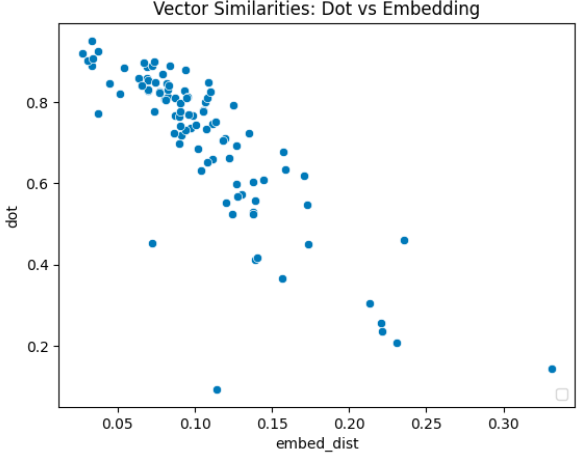}
\caption{\label{fig:6.1_ScatterDotvEmbed} Comparison of Vector Similarities of Dot Score and Embedding Distance}
\end{figure}

\begin{figure}[!h]
\centering
\includegraphics[width=.48\textwidth]{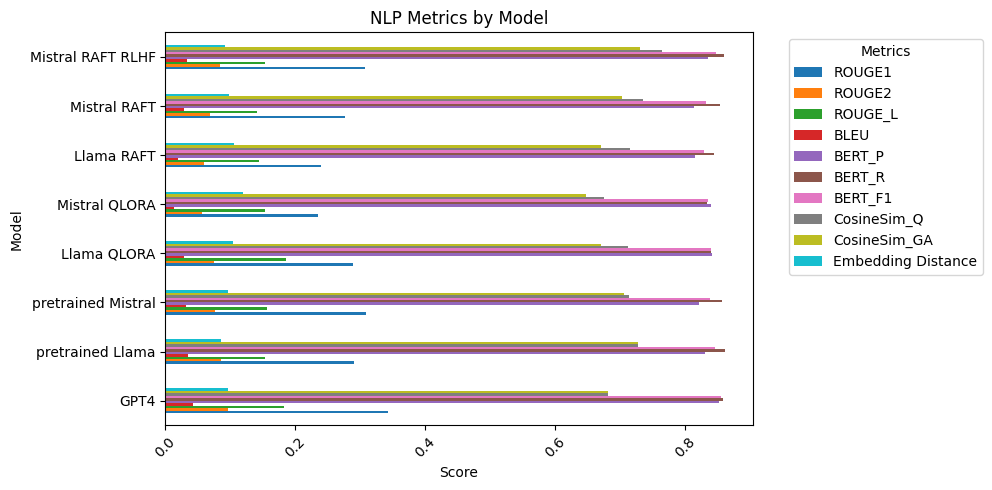}
\caption{\label{fig:nlp bar} Metrics Subset: Traditional NLP Comparisons}
\end{figure}

Of quantitative metrics, Figure \ref{fig:6.1 metrics with variance} shows only two quantitative metrics had variance: GPT-4 dot score and golden answer dot score. GPT-4 dot score found Mistral RAFT RLHF as best and LLaMa QLoRA as the worst. Golden answer dot score was most aligned with pretrained LLaMa, Mistral RAFT as next best, LLaMa QLoRA as the worst. There were more dot score comparisons that did not have variance, shown in Figure \ref{fig:dot bar}. Figure \ref{fig:6.1_ScatterDotvEmbed} shows a negative correlation, as expected, with dot score and embedding distance on a scatter plot. The traditional NLP metrics had no variance, but are visualized in Figure \ref{fig:nlp bar}. The bar graph distribution is nearly the same across all models.


\begin{figure*}[!h]
\centering
\includegraphics[width=.8\textwidth]{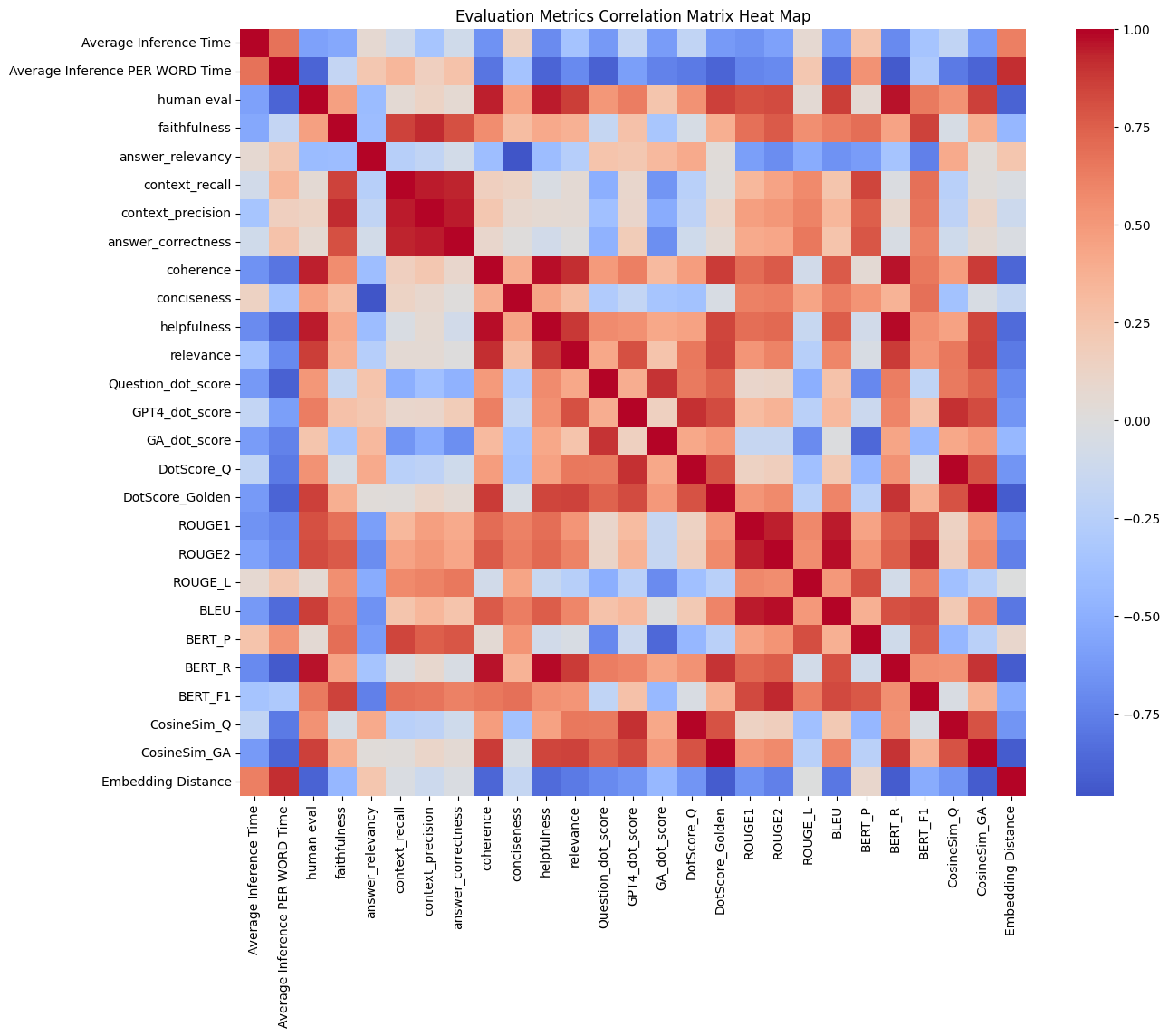}
\caption{\label{fig:correlation_matrix_heatmap} Metrics Correlation Matrix Heatmap}
\end{figure*}

Figure \ref{fig:correlation_matrix_heatmap} shows all the evaluation metrics as a correlation matrix heatmap with red representing a positive correlation and blue representing a negative correlation. Given the disparity of Ragas metrics with human evaluation, we are particularly interested in analyzing the correlations with human evaluation, which has a positive correlation with the following metrics, OpenAI GPT-4 (coherence, consciousness, helpfulness, relevance), and quantitative metrics with no variances: dot scores, ROUGE1, ROUGE2, BLEU, BERT\_R, BERT\_F1, cosine similarity, and embedding distance. This indicates that OpenAI GPT-4 metrics are most aligned with human evaluation, whereas Ragas metrics are not. While there were many positive correlations with quantitative metrics, there was no variance, indicating these metrics are too simplistic to fully evaluate the complexity of LLMs inference.

\subsection{RLHF Analysis}

An RLHF training pipeline for LLMs is done in three main parts: domain-specific fine-tuning, supervised instruction fine-tuning (SFT), and reward modeling in RLHF.\cite{iyer2023} Domain-specific fine-tuning contains domain-specific data only, which will only contain conversational non-supervised Reddit data with the answers only. Thus, the input is domain knowledge. Then, the second part of the RLHF training pipeline is supervised fine tuning where there is a target label and we fine-tune the travel-domain LLM that is specified on certain tasks and domain-specific pairs such as Q\&A pairs, prompts and instructions, and responses curated for our travel domain. The output of domain specific pretraining is a model that is able to recognize the context from input and be able to predict the next words/sentences. So, we perform SFT with prompt-text pairs (Q\&A pairs) to provide the pretrained LLM with knowledge, specifically travel-domain specific tasks. Thus, it will be able to respond to context-specific questions curated to our specified travel domain. The output of this second phase is a LLM that mimics a conversational agent. The final phase is training to perform RLHF utilizing human feedback on a LLM, so reward model training trains the LLM to classify responses as being good or bad based on a thumbs up or thumbs down icon of binary range of 0 or 1, or face emojis that range from 0.0, 0.25, 0.50, 0.75, and 1.0. The RLHF training pipeline is particularly useful to prevent biased responses from responses generated from training and at inference time, thus, this pipeline can obtain desirable and accurate responses from LLMs and helps with domain-specific LLMs as well. However, it is important to note that human involvement is needed to manually score and screen for responses.\cite{iyer2023}

\begin{figure}[!h]
\centering
\includegraphics[width=.45\textwidth]{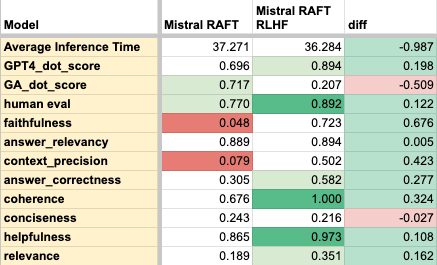}
\caption{\label{fig:6.1rlhf}Non-zero variance evaluation metrics for Mistral RAFT before and after RLHF}
\end{figure}

Figure \ref{fig:6.1rlhf} shows the nonzero variance metrics before and after RLHF for the best model Mistral RAFT. We expect RLHF to improve the performance, so before was subtracted from the after, and we expect to see positive (green) numbers in the difference column, although this is reversed for metrics (i.e. runtime) where lower is better, then we would expect to see negative (also green) numbers. We see that RLHF performance better than its pretrained counterpart in all metrics except golden answer dot score and conciseness. RAFT is known to give lengthy answers with multiple answer options, so it is interesting that RLHF further increased this aspect. Conciseness would also impact the dot score as the vector length is assumed to be much longer than the golden answer to which it is being compared.

\subsection{Inference Generation Evaluation}
Tools used to generate inference include a range of quality of GPUs, which affected inference speed. The preferred method to generate inferences was to host on HuggingFace, deploy to AWS, generate inferences via an inference endpoint, and pay the nominal fees associated. The processing speed was ten times faster than generating inferences on the GPUs with which the models were trained. 

During experimentation there were constraints when evaluating our models as certain models have a limited maximum context window compared to others in OpenAI, such as GPT-3.5 Turbo versus GPT-4 Turbo Preview. There are limitations in terms of capacities of the models usage. If the input prompt is too long the token size is larger, thus longer time is needed for evaluation, which impacts evaluation performance. The input prompts were particularly long, which is expected for Reddit data. There are also limited query requests to the API. Gpt-4 turbo preview with a context window size of 128,000 tokens and a API usage rate limit of 450,000 tokens per minute versus GPT-3.5 turbo with 16,385 tokens and a rate limit of 80,000 tokens per minute \cite{openaiplatform_2024}. Thus, upgrading from GPT-3.5 turbo to GPT-4 turbo preview was needed to send 33 questions with 10 metrics to OpenAI for evaluation for all models.

The deployed model on HuggingFace runs exceptionally faster than all of the local models with a median latency of 3 seconds. At the 95th percentile of all 39 requests to date since the model was deployed, 4.80 seconds is one of the shortest inference times compared to local models which may consume up between 2 to 3 minutes on average to complete. When connected to the inference endpoint NVIDIA A100 GPU on HuggingFace, it costs a dollar per hour. Overall, generating inferences cost about 10 dollars, less than 10 hours, and significantly faster than several days of nonstop computing in our prior experience.

\begin{figure}[!h]
\centering
\includegraphics[width=.48\textwidth]{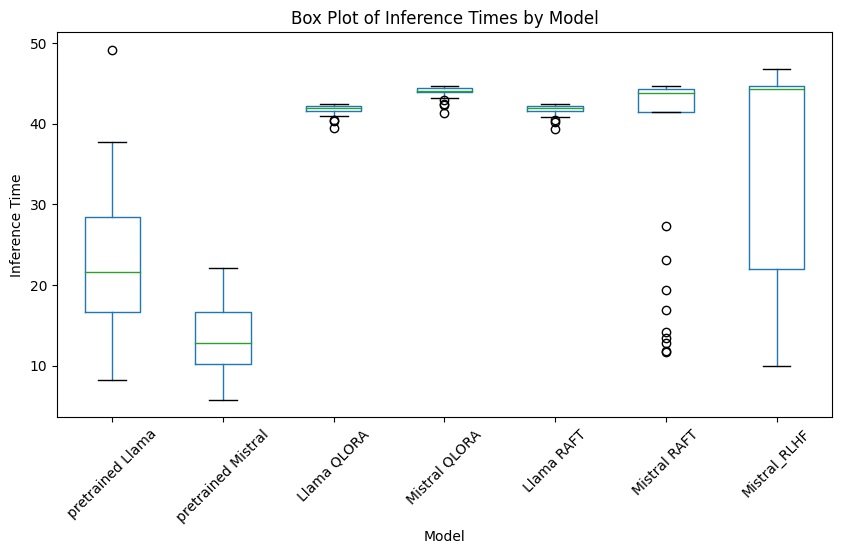}
\caption{\label{fig:box_plot_inference_times} Distribution for simulated runtimes across all fine-tuned and baseline LLMs }
\end{figure}

Average inference times ranged from 14 to 44 seconds (see Figure \ref{fig:box_plot_inference_times}). This figure also shows little variance in inference time amongst LLaMa QLoRA, Mistral QLoRA, and LLaMa RAFT. Mistral RAFT has little variance as well, but several outliers. Pretrained LLaMa and Mistral and Mistral RLHF had a much larger variance in inference time, with pretrained Mistral clearly as the fastest. This shows the models are not always consistent when generating responses, as it depends on the context, prompt, and complexity, and length of the question as well. For context of the prior method of generating inferences, pretrained Mistral went from an average of 3008 seconds to 14 seconds, slowest to fastest.

\section{Conclusion}
The travel industry currently lacks chatbots that can deliver real-time insights for all questions and concerns related to itinerary planning, safety recommendations, and tips for navigating as a tourist. With the onset of LLMs, the time required to construct an artificial assistant has been accelerated and the potential for building tailored chatbots is becoming more apparent. However, deriving value from new technological advancements with LLMs requires thorough research into proposed techniques and comprehensive evaluations to gauge their utility and usefulness as viable fit-for-purpose tools. To meet the demand for a fit-for-purpose chatbot, various implementation strategies for fine-tuning foundational models and evaluation metrics are explored and evaluated through a series of experimentation.

This research compared two fine-tuning methods (QLoRA and RAFT) on foundational LLMs (Mistral and LLaMa 2 7B), finding that Mistral RAFT outperformed the other three combinations of those models: 1) Mistral QLoRA, 2) LLaMa RAFT, and 3) LLaMa QLoRA. Based on this initial assessment, a final fine-tuning process with Reinforcement Learning with Human Feedback is adapted to further upskill models of interest. Mistral RAFT was further fine-tuned with RLHF and outperformed all models including baseline models (GPT-4, pretrained LLaMa, pretrained Mistral). This refining process yielded an optimal model trained to recognize inputs and contexts in order to deliver inferences aligning with our expectations. Promoting our fine-tuned model that was trained on the RAFT dataset and human feedback as an inference endpoint.

This research also compared LLM evaluation metrics such as the End-to-End (E2E) benchmark method of "Golden Answers," traditional natural language processing (NLP) metrics, RAG Assessment (Ragas), OpenAI GPT-4 evaluation metrics, and human evaluation. An evaluation set with 37 questions and “Golden Answers” was formulated synthetically and manually to ensure a diverse set of performances against 4 of our models and 3 baseline models (see Appendix \label{app:AppendixD}). While Mistral RAFT RLHF performed best on human evaluation, it performed the worst for some Ragas metrics. Mistral QLoRA performed the worst on most of the quantitative metrics, best on most of the Ragas metrics.

Here are the key findings from our research: 
\begin{itemize} 
\item Quantitative and Ragas metrics do not align with human evaluation.
\item OpenAI GPT-4 evaluation metrics most closely align with human evaluation.
\item It is essential to keep humans in the loop for evaluation, as traditional NLP metrics are insufficient.
\item Mistral generally outperforms LLaMa.
\item RAFT outperforms QLoRA but still requires post-processing.
\item RLHF improves model performance significantly.
\end{itemize}

\begin{figure*}[!h]
\centering
\includegraphics[width=.85\textwidth]{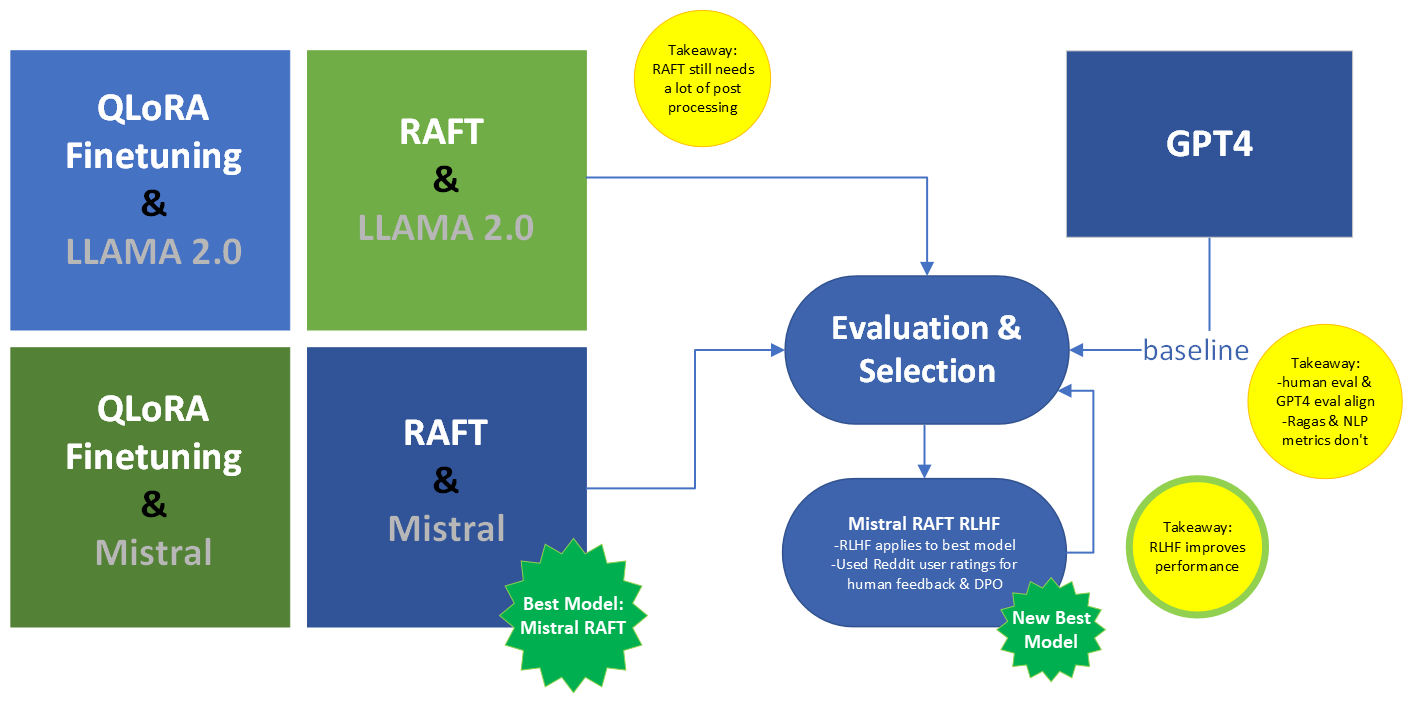}
\caption{\label{fig:7final_overview}Project Overview with Results and Takeaways}
\end{figure*}

To conduct this research, a travel dataset was sourced from the Reddit API by querying travel-related subreddits to gather travel conversation prompts and personalized travel experiences, and then augmented for each fine-tuning method including formats like Q\&A format, RAFT, and RLHF. These datasets are available for further research. As part of the data cleaning process, a dot score, typically used for quantifying semantic similarity, was employed to potentially reduce the amount of noise in raw datasets. By iteratively transforming for higher quality inputs, the model’s ability to learn and generalize improved. Much of the experimentation involved fine-tuning with QLoRA where preferred examples for probable questions and responses were provided to the model. Ultimately through successive iterations, 5 fine-tuned models: 1) LLaMa 2 QLoRA, 2) Mistral QLoRA, 3) LLaMa 2 RAFT, 4) Mistral RAFT, 5) Mistral RAFT RLHF were publicly deployed on HuggingFace (see Figure \ref{fig:7final_overview}).

\subsection{Discussion}
As inputs for the two RAFT models, this novel technique was successfully implemented, and a new synthetic RAFT dataset was augmented from knowledge, originally intended for Q\&A format for QLoRA, that integrates positive with negative documents and a chain-of-thought response manner to promote logical reasoning. Utilizing the new approach called RAFT, we were able to input domain knowledge to be baked in from the model weights, be able to read in information from retrieved results, and ultimately allowed RAFT LLMs to outperform QLoRA. Thus, this enables smaller models like Mistral and LLaMa 2 7B parameter models to save on inference cost. This is shown in our load and response times in the next section as RAFT was able to train and process much quicker. For example, for LLaMa 2 7B, it took around 33 minutes to train for 10 epochs and 20 steps. A constraint of RAFT is the need to have a very diverse set of question and answer pairs and randomization of ground truth documents are required to help improve performance.

Fine-tuning is a sensitive process and suboptimal input training sets may run the risk of disrupting the foundational model’s inherent ability to process inputs and outputs efficiently. Despite cleansing and transformation procedures to ensure acceptable data quality standards, there is still potential noise that is not conducive to model fine-tuning. Although our final best model was best rated by humans, the models stand to improve the most from better data quality and quantity. Aside from critical dependencies on hardware options to run time-consuming training processes, there is a recognized complexity with tuning hyperparameters with respect to batch sizing, epoch amounts, learning rates, gradient steps and other components of the architecture.

Currently, the models may produce a lot of noise and there is limited control over how inputs are processed and embedded. Thus, it is dependent on the post-processing to parse out unrelated and irrelevant segments that are needed to achieve human-readable outputs. Due to budget and time constraints, comprehensive prompt tuning was not explored and it could be hypothesized that bespoke prompt templates for identified tasks can be pre-written to guide model predictions.

Despite the shortcomings, there remains much potential for mining insights from real-world conversations taking place on online web forums or social media platforms. While pretrained and foundational models have been conditioned to not express any toxicity in their responses, it can still learn and adapt from unfiltered contexts, such as children as they become more understanding of the actual world outside of their homes. As open-source tooling continues to evolve in capability, the boundaries for possibilities will scale exponentially. Frameworks such as HuggingFace are expanding the limits when it comes to fine-tuning with no-code, low-code solutions to rapidly train numerous models in parallel. It is becoming feasible to transform input datasets into preferred formats for model interpretation.

As hardware continues to evolve, the expenses related to hosting and serving model inference endpoints should reduce and save financial room for investing in richer knowledge bases. Updating the knowledge base would be informed by measuring model performance degradation over time, especially if users are consistently asking about particularly niche topics that are not concurrent with the knowledge base. If there are noticeable disparities, the knowledge base would be reviewed for outdated information and imputed for new ones that better align with user expectations.

Furthermore, to ensure model responses are aligned with human expectations, extensive RLHF training with DPO using large-scale preference datasets can also be considered. Since inferences have been generated for all 5 model combinations, the ideal and suboptimal model responses can be leveraged as examples for further response adjustments.

Sourcing out relevant training data and ensuring strict adherence to quality standards is a persistent challenge for supplying LLMs with in-domain knowledge. Transforming raw and unstructured corpora of texts into preferred formats requires a generalizable pipeline from ingestion to loading. Though use-case dependent, once the texts are cleaned, they must be chunked in an optimal manner that preserves semantic integrity. In order to achieve the numerical representation of texts that models can understand, the text must be embedded and it is evident that different embedding models may yield different results. 

Processing with LLMs requires available resources both for fine-tuning and inference generation. While the data collection, transformation, and processing steps can be performed with local hardware, there is much benefit gained from utilizing powerful resources to accelerate the data extraction and ingestion process. For large corpora of textual data, the time required to fine-tune in an optimal manner is extensive. Without physical access to GPUs, the costs related to training via virtual GPUs can also scale quickly. 

The landscape for AI advancement is changing rapidly and techniques that may be model or SOTA today may become obsolete in the near future as operational efficiency with running, training, and evaluating LLMs evolves. By exploring the foundational progress as early-adopters, we can better understand and appreciate the coming offerings of new researched methodologies.

\subsection{Future Work}
There remains a wealth of unanswered questions about the prospects for Large Language Models. First and foremost, with data being the differentiating factor between good and bad performances, it is crucial that training sets are in strict alignment with human expectations and preferences. Future data collection processes might involve additional data sources that can provide more insights about particular destinations in conversational form to further fine-tune the model's ability to generalize in some intended manner. It would be interesting to nichely fine-tune the model to a specific city or country so data collection efforts can be more focused with real time webscraping of relevant website.

Considering the cutoff points for the pretrained models, it is also possible to implement continuation of in-domain pre-training by fine-tuning for next token prediction. This would mean updating or expanding the model’s parameters to integrate update knowledge. An alternative to fine-tuning would be advanced federated RAG systems that models can access and query from depending on the requirement from user queries. The RAG systems would be representative of the scope of the chatbot’s deliverables and a retrieval mechanism would be tailored to the nature and distribution of the contents within the knowledge banks.

Implementation-wise, there are a few considerations that are yet to be explored, such as fine-tuning embeddings models to better represent raw or cleaned texts numerically. The process involving mapping texts to numbers leverages default techniques with open-source embeddings models that may not be entirely fit-for-purpose. Instead, prior to loading a database with embedded vectors, it would be worthwhile to experiment and evaluate with various in-domain embeddings models.

\bibliographystyle{IEEEtran}
\bibliography{DATA298.bib}

\onecolumn
\begin{appendices}
\begin{landscape}

\section{External Links}
\label{app:AppendixA}

All source code generated for the project is located in the group's GitHub repository: 
\href{https://github.com/soniawmeyer/WanderChat}{https://github.com/soniawmeyer/WanderChat}

Golden questions and answers used for evaluation are available here: \href{https://docs.google.com/spreadsheets/d/1kJdUUte8bJWywCvmn91CsRua0Kn7BOftcRRozE4YH00/edit?usp=sharing}{GQ GA}

Travel curated datasets for fine-tuning are available on HuggingFace: 
\begin{itemize}
    \item \href{https://huggingface.co/datasets/soniawmeyer/reddit-travel-QA-fine-tuning}{https://huggingface.co/datasets/soniawmeyer/reddit-travel-QA-fine-tuning}
    \item \href{https://huggingface.co/datasets/soniawmeyer/conversations-filtered-travel}{https://huggingface.co/datasets/soniawmeyer/conversations-filtered-travel}
\end{itemize}

\begin{table}[h!]
\caption{fine-tuned Models Hosted on Hugging Face}
\centering
\begin{tabular}{|>{\raggedright\arraybackslash}p{3cm}|>{\raggedright\arraybackslash}p{10cm}|}
 \hline
 \textbf{Model} & \textbf{URL} \\
 \hline
 Mistral QLoRA & \url{https://huggingface.co/sherrys/mistral-2-7b_qlora_falcon_426/tree/main} \\
 \hline
 LLaMa QLoRA & \url{https://huggingface.co/beraht/LLaMa-2-7b_qlora_falcon_417} \\
 \hline
 Mistral RAFT & \url{https://huggingface.co/sherrys/426_mistral_RAFT_50e_10s} \\
 \hline
 LLaMa RAFT & \url{https://huggingface.co/beraht/LLaMa2_Falcon_RAFT_50e_10s/tree/main} \\
 \hline
 Mistral RAFT RLHF & \url{https://huggingface.co/chriztopherton/Wanderchat_mistral_RAFT_RLHF} \\
 \hline
\end{tabular}
\end{table}

\section{Complete Evaluation Metrics}
\label{app:AppendixD}
\begin{figure}[!h]
\centering
\includegraphics[width=\linewidth]{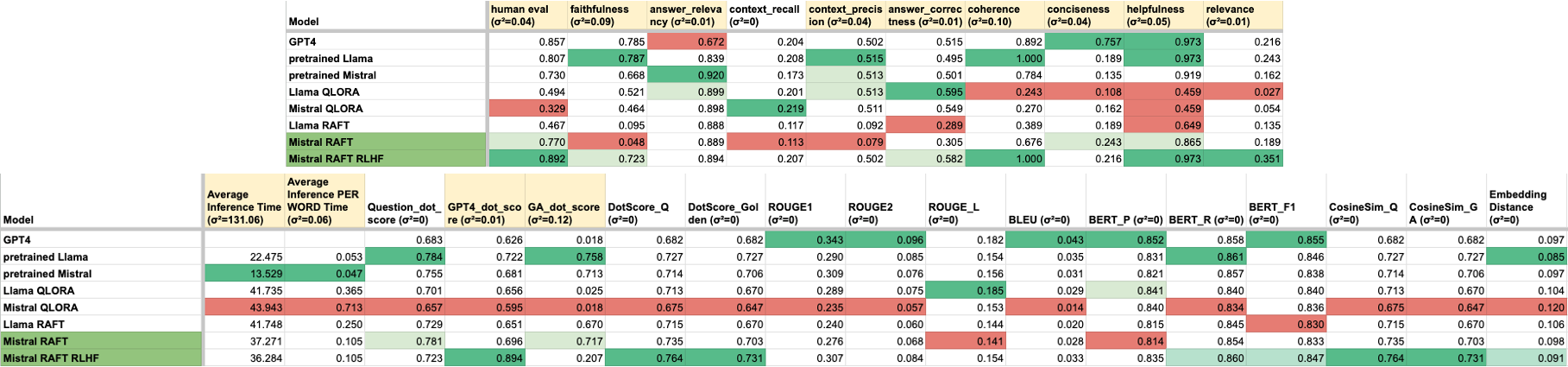}
\caption{Complete metrics for all models, qualitative on top, quantitative on bottom, non zero variance metrics highlighted in tan, best (and next best if best is a baseline model) and worst performing models highlighted in green and red respectively}
\end{figure}
\end{landscape}

\end{appendices}

\end{document}